\documentclass[10pt, twocolumn, letterpaper]{article}

\usepackage[pagenumbers]{cvpr} 

\usepackage{amsfonts}
\usepackage{amsmath}
\usepackage{graphicx}
\usepackage{threeparttable}
\usepackage{multirow}
\usepackage{makecell}
\usepackage{booktabs}
\usepackage{colortbl}

\newcommand{\lred}{\color{red!60!black}}
\newcommand{\lgreen}{\color{green!50!black}}
\newcommand{\lblue}{\color{blue!60!black}}

\definecolor{cvprblue}{rgb}{0.21,0.49,0.74}
\usepackage[pagebackref, breaklinks, colorlinks, allcolors=cvprblue]{hyperref}

\def\paperID{16502} 
\def\confName{CVPR}
\def\confYear{2025}

\title{Image Quality Assessment: Investigating Causal Perceptual Effects with Abductive Counterfactual Inference}

\author{
	Wenhao Shen, 
	Mingliang Zhou, 
	Yu Chen, 
	Xuekai Wei, 
	Jun Luo, 
    Huayan Pu, 
	Weijia Jia
}

\graphicspath{{figure/}}

\begin{document}
\maketitle

\begin{abstract}
Existing full-reference image quality assessment (FR-IQA) methods often fail to capture the complex causal mechanisms that underlie human perceptual responses to image distortions, limiting their ability to generalize across diverse scenarios. In this paper, we propose an FR-IQA method based on abductive counterfactual inference to investigate the causal relationships between deep network features and perceptual distortions. First, we explore the causal effects of deep features on perception and integrate causal reasoning with feature comparison, constructing a model that effectively handles complex distortion types across different IQA scenarios. Second, the analysis of the perceptual causal correlations of our proposed method is independent of the backbone architecture and thus can be applied to a variety of deep networks. Through abductive counterfactual experiments, we validate the proposed causal relationships, confirming the model's superior perceptual relevance and interpretability of quality scores. The experimental results demonstrate the robustness and effectiveness of the method, providing competitive quality predictions across multiple benchmarks. The source code is available at https://anonymous.4open.science/r/DeepCausalQuality-25BC.
\end{abstract}

\section{Introduction}
Image quality assessment (IQA) is a fundamental area within image processing that is dedicated to evaluating the perceptual quality of images.
The primary objective of IQA is to quantify how closely an image aligns with the human visual system (HVS), particularly when subjected to various distortions such as compression, noise, or blurring.
Given its importance, IQA is integral to numerous applications, including image compression, transmission, and enhancement \cite{CKDN, Chen_TIP_2021, ding_2021_IJCV}, where preserving visual fidelity is critical.
IQA methods are generally classified into three main categories: full-reference, reduced-reference, and no-reference approaches, all of which can be considered typical regression problems \cite{DeepQA, Zhou_TBC_2022, Lan_TBC_2023}.

In full-reference image quality assessment (FR-IQA), the evaluation process is typically divided into three key stages: feature decomposition, feature comparison, and perceptual score mapping.
Recent advancements have highlighted the pivotal role that deep learning methods play in aligning IQA predictions with human subjective scores \cite{DeepSim, AHIQ}.
Notably, deep representations of images have been shown to strongly correlate with perceptual relevance, emphasizing their significance in quality assessment \cite{LPIPS, DISTS, Shen_TIP_2024}.
In deep networks, different channels capture distinct features of original and distorted images, collectively encoding perceptual information.
When quality degradation occurs, these features in the distorted image become contaminated to varying degrees, resulting in altered distinctions from those of the reference image \cite{Liao_2024_TIP}.
However, the mere alteration of features is insufficient to fully capture the perceptual quality differences between the original and distorted images, as changes in correlation alone do not adequately explain the complex, nonlinear effects of distortions on human visual perception.
Furthermore, while existing methods are effective in capturing statistical dependencies, they fail to reflect the intrinsic mechanisms that drive changes in perceived quality.
Specifically, these methods only quantify how distortions affect the similarity between feature representations and cannot elucidate how these distortions influence human perception. Consequently, they are unable to differentiate between distortions that may exhibit similar feature distances yet have significantly different perceptual impacts. This limitation restricts their capacity to provide accurate quality assessments.
Additionally, no existing theoretical framework fully explains why IQA remains such a challenging task or why certain methods succeed in some contexts and fail in others.
This lack of clarity impedes advancements within the field.

To address these challenges, we define IQA as a counterfactual inference problem, offering a perspective that bridges existing theoretical gaps \cite{Shams_2010_TCS}.
Counterfactual reasoning provides a formalized structure for defining the minimal perceptual cost, enabling a more precise understanding of how distortions affect perceived quality.
Given that the reference image serves as the baseline for the original visual content, an effective IQA method should be able to predict scores on the basis of the perceptual distortions present in degraded images. However, the manner in which distortions are perceived is essentially probabilistic and is influenced by a variety of contextual factors and prior experience.
While deep learning-based methods are adept at capturing complex feature interactions, they often fail to incorporate the probabilistic relationships that underlie perceptual judgments.
By constructing IQA from a causal reasoning perspective, our approach aims to quantify perceptual differences in a way that reflects these inherent uncertainties and complexities.
This causal perspective provides us with not only a measure of feature similarity but also a more nuanced understanding of how different distortions affect human perception.
The main contributions of our study are as follows:
\begin{itemize}
	\item
	We investigate the causal effects of deep network features on perception, effectively integrating causal relationships with feature comparisons to construct a deep network-based FR-IQA model.
	By employing distance comparisons among the causal relationships of deep features, the model can accurately address complex distortion types encountered across various image quality assessment scenarios, resulting in competitive quality prediction outcomes.

	\item
	The proposed method exhibits significant generality, as the analysis of perceptual causal correlations within deep features is independent of the backbone architecture.
	This characteristic enables the implementation of the approach across different deep backbone networks, thereby enhancing its versatility and applicability in diverse contexts of image quality assessment.

	\item
	The validity of the proposed causal relationships is substantiated through experimental evidence derived from abductive counterfactual inference, which demonstrates superior perceptual relevance and score interpretability and establishes its robustness and practicality.
\end{itemize}

\section{Related Work}

\subsection{Deep Network-based FR-IQA}
Deep networks have been employed to model the complex relationships between reference and distorted images in FR-IQA.
In \cite{DeepQA}, convolutional operations on both reference and distorted images are utilized to obtain sensitivity maps, which are then applied to the difference map for quality estimation.
Gao \textit{et al.} \cite{DeepSim} measured the local similarity between features extracted from deep neural networks (DNNs) for both reference and distorted images. The local quality indices are aggregated to produce the overall quality score.
Subsequent studies have shown that VGG network features serve as effective training losses for tasks such as image synthesis.
Notably, the study of \cite{LPIPS} compares deep features across various architectures and tasks, showing that perceptual similarity is an emerging property shared by deep visual representations.
Hence, some advancements focus on perceptual similarity or differences in deep feature space.
For example, Ding \textit{et al.} \cite{DISTS} employed the structural similarity index (SSIM) between deep features to measure perceptual texture similarity, offering a unified deep image structure and texture similarity (DISTS) framework, and the adaptive DISTS (ADISTS) \cite{ADISTS} improved it by using a dispersion index to identify texture regions and adaptively aggregate structure and texture measurements.
In contrast, Liao \textit{et al.} \cite{DeepWSD} projected image degradations into perceptual space and computed the Wasserstein distance in deep feature space, highlighting the importance of distributional differences in deep feature representations for quality assessment.

Existing methods are limited in capturing perceptual quality differences because they rely on feature alterations and statistical dependencies and fail to model the nonlinear effects of distortions on human perception.
They quantify feature similarity without revealing how distortions impact perception, leading to difficulties in distinguishing distortions with similar feature distances but different perceptual effects.
Additionally, the absence of a robust theoretical framework to explain why some methods succeed or fail in different scenarios limits progress in IQA.

\subsection{Causal Inference in Deep Learning}
In deep learning, with the presence of confounding factors, the network can easily capture spurious correlations between inputs and their outputs.
Therefore, the application of causal inference integration in deep learning has received increasing attention.
Causal inference provides a theoretical foundation for disentangling correlation from causation, which is particularly useful for improving model generalizability, robustness, and interpretability \cite{Kim_2023_CVPR} \cite{Yang_2023_KDD}.
In deep learning, causal approaches have been explored to mitigate issues such as bias, confounding, and spurious correlations that often arise from data-driven methods \cite{Cai_2023_NEURIPS, Song_2024_CVPR}.
Yue \textit{et al.} \cite{Yue_2020_NEURIPS} proposed an interventional few-shot learning method that applies backdoor adjustment to mitigate the confounding effect of pretrained knowledge.
Liu \textit{et al.} \cite{Liu_2022_CVPR} addressed both visual confounders in the object detection stage and linguistic confounders in the decoding process for image captioning tasks via causal intervention, leading to substantial improvements in generalizability across different datasets.
Hu \textit{et al.} \cite{Hu_2023_TKDE} modelled the causal relationship between image-text matching and fake news detection in multimodal settings and used causal inference to eliminate confounding effects during training and leverage matching bias to improve detection accuracy.

In the context of image quality assessment, causal reasoning offers a promising avenue for improving the accuracy and consistency of quality prediction.
By focusing on the causal relationship between image features and perceived quality, it can explore how causal models mitigate the problems of spurious correlation and overfitting to better align with human perceptual judgments and improve the reliability of deep learning models in real-world applications.

\begin{figure*}[t]
	\centering
	\includegraphics[width=1\textwidth]{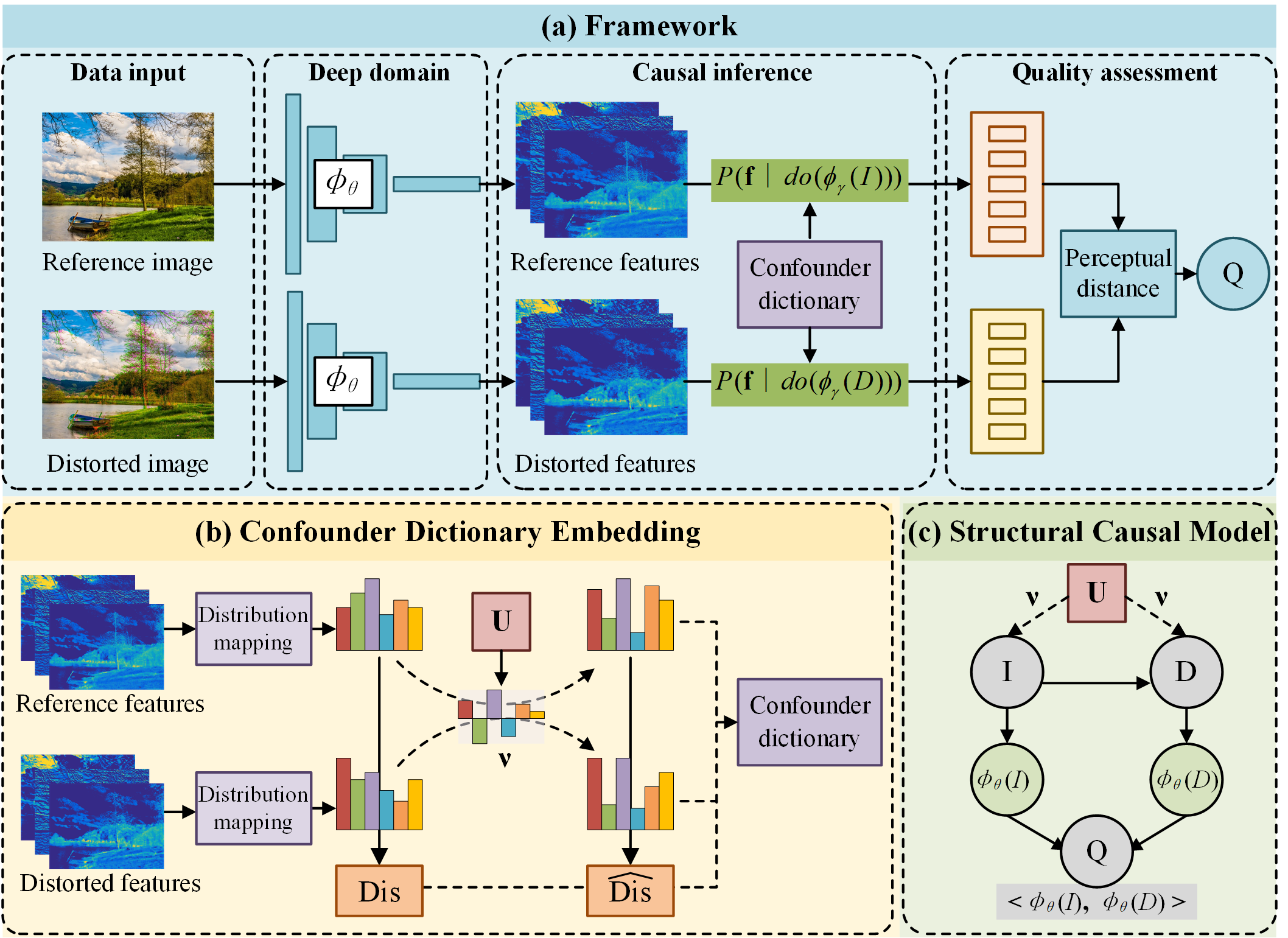}
	\caption{
		The framework of our proposed method.
		Details of the confounder dictionary embedding and the structural causal model are detailed below.
	}
	\label{fig:framework}
\end{figure*}

\section{Problem Formulation}

The general objective in image quality assessment is to minimize the discrepancy between the predicted quality score and the subjective mean opinion score (MOS), which can be formalized as:
\begin{equation}
	\min_{\zeta, \omega, \theta} \mathbb{E} [l(f_\zeta(\operatorname{Dis}_\omega(\phi_\theta(I), \phi_\theta(D))), M O S)]
\end{equation}
where $\theta$ represents the parameters of model $\phi$, $\operatorname{Dis}(\phi_\theta(I), \phi_\theta(D))$ denotes the distance between the feature representations of the original image $I$ and the distorted image $D$ in the deep feature space $\phi(\cdot)$, and $l(\cdot)$ is the loss function that quantifies the difference between the predicted score $f_\zeta(\operatorname{Dis}_\omega(\phi(I), \phi(D)))$ and the subjective quality.
The expectation $\mathbb{E}$ is taken over the data distribution.

For full-reference image quality assessment, we consider the causal relationship between the original image $I$ and the distorted image $D$.
The relationship can be expressed as follows:
\begin{equation}
	\mathcal{Q}_S = m(\phi_\gamma(I, D)), \quad \eta \perp \gamma \text{ and } \gamma,\eta \sim F_\theta
\end{equation}
Here, $m(\cdot)$ represents a function that maps the joint feature representation $\phi_\gamma(I, D)$ of the original and distorted images to the subjective quality $\mathcal{Q}_S$.
The parameter $\gamma$ is pivotal in this formulation, as it represents the features with causal implications for image quality, thereby influencing the perceptual evaluation.
In contrast, $\eta$ signifies the noise components that are independent of the deep representation parameter $\gamma$, suggesting that these components do not contribute to the quality assessment.
Both $\gamma$ and $\eta$ are derived from the distribution $F_\theta$, indicating their dependence on the model's parameters, $\theta$.
Specifically, $\theta$ represents the foundational parameters of the deep feature representation, which gives rise to two distinct components: $\gamma$, the relevant features that causally affect image quality, and $\eta$, the extraneous noise that does not contribute to the quality assessment.
This causal relationship isolates the effect of distortions from external factors, ensuring that the predicted quality score is influenced primarily by distortions between the reference and distorted images, whereas other factors are captured by $\eta$.

In light of the established causal relationship, the subsequent phase focuses on prioritizing causal features $\gamma$ while effectively disregarding the noise components $\eta$.
The goal is to adjust the model parameters $\theta$ to better capture the relevant causal features $\gamma$ that significantly influence the subjective quality score $\mathcal{Q}_S$.
The optimization process is designed to minimize the prediction error across all conceivable data distributions, thereby ensuring that the model remains robust against varying inputs and distortions.
This objective can be formally articulated as follows:
\begin{equation}
	\min_{\zeta, \omega, \theta \to \gamma} \sup_{P \in \mathcal{P}} \mathbb{E}_{P} [l(f_\zeta(\operatorname{Dis_\omega}(\phi_\theta(I), \phi_\theta(D))), MOS)]
\end{equation}
where the term $\min_{\theta \to \gamma}$ signifies the optimization of model parameters $\theta$ to enhance the representation of the causal parameters $\gamma$,
$\mathcal{P}$ represents the set of all possible distributions that are consistent with the structural causal model, and the supremum operator $\sup_{P \in \mathcal{P}}$ accounts for the worst-case scenario over these distributions.

In our study, instead of training the model from scratch, we define $\theta$ as the pretrained model weights.
We subsequently employ variable interventions to validate the autoregressive properties of the model, facilitating the filtration of extraneous information.
This strategy aims to consider only causal representation parameters $\gamma$ in the quality prediction process while effectively ignoring the noise component $\eta$.

\section{Methodology}

\subsection{Framework}
As illustrated in \autoref{fig:framework} (a), the proposed framework integrates multiple stages, starting with the input of reference and distorted images.
These images are first processed in the deep domain, where deep learning models extract high-level feature representations.
These feature representations serve as the foundation for the subsequent causal inference stage.
In this stage, the relationships between the extracted features are analysed to identify causal variables that are robust to distortions and potential confounders.
Finally, the quality assessment step uses these identified causal variables to evaluate the perceptual quality of the distorted image, producing a quality score that reflects the minimal perceptual cost on the basis of the underlying causal structure.

To explain the causal relationships considered in our approach, we employ a structural causal model (SCM) for the full-reference image quality assessment.
The SCM represents variables as nodes and relationships as edges in a graphical structure, using directed arrows to indicate causal relationships between two variables.
In our model, we incorporate the reference image $I$, the distorted image $D$, the representations of both images $\phi_\theta(I)$ and $\phi_\theta(D)$, and the image quality score $Q$, as depicted in \autoref{fig:framework} (c).
We assume that the images are characterized by factors that have causal relationships with quality, along with elements that are unrelated to quality.
In this framework, image quality is directly determined by factors that causally influence it.
Simply using a training process or pretrained weights from other fields to assess image quality typically yields representations that are correlated with perceptual quality but do not capture causal relationships.
Therefore, we introduce an unknown exogenous variable $U$ to account for uncertainties and identify the representations that are causally related to quality.

\subsection{Deep Causal Measurement}
In our approach, to verify causal relationships in perceptual quality assessment, we design a causal validation process using an SCM with an exogenous variable $U$.
This process begins with a causal intervention to test the effect of image characteristics on perceived quality, using the operator $\text{do}(\cdot)$ to map both reference and distorted images onto a shared causal feature space. The intervention can be formulated as follows:
\begin{equation}
	\mathbf{f}_I = \phi_\theta(\text{do}(I = i)), \quad \mathbf{f}_D = \phi_\theta(\text{do}(D = d))
\end{equation}
where $\phi_\theta$ is the feature extraction function parameterized by $\theta$, which maps the images into the deep feature space, and $\mathbf{f}$ denotes the deep features.
To substantiate the causal nature of the features, we introduce the exogenous variable $U$ and perform simultaneous interventions $v$ on both the original reference image features $\phi_\theta(I)$ and distorted image features $\phi_\theta(D)$ to generate the intervened causal representations $\mathbf{f}'$.
This intervention yields representations formulated as follows:
\begin{equation}
	\mathbf{f}'_I = \phi_\theta(\text{do}(I, v)), \quad \mathbf{f}'_D = \phi_\theta(\text{do}(D, v))
\end{equation}
Next, we calculate the perceptual distance $\operatorname{Dis}(\cdot)$ before and after the intervention to assess the invariance of the causal representation. If the perceptual distance differs significantly between the pre- and postintervention states, then the feature representation can be deemed causal, as it is sensitive to distortions influencing perceptual quality. This invariance can be evaluated as follows:
\begin{equation}
	\Delta = \operatorname{Dis}(\mathbf{f}_I, \mathbf{f}_D) - \operatorname{Dis}(\mathbf{f}'_I, \mathbf{f}'_D)
\end{equation}
If $\Delta \neq 0$, it confirms that the causal factors within the feature representations significantly affect perceptual quality.
In practice, we vary the disturbance intensity to determine the strength of the perceptual causal relationship within the feature representations, and we use a confounder dictionary to record the representations that maintain causality at all intensities, denoted $\Gamma(\mathbf{f}_I, \mathbf{f}_D)$.
The causal influence on perceptual quality is then quantified by observing how the perceptual distance changes with varying intervention intensities.

The regression invariance is validated if the perceptual distance $\operatorname{Dis}(\mathbf{f}'_I, \mathbf{f}'_D)$ yields a consistent quality score $Q$ under intervention, where $\mathbf{f}$ is parameterized by $\gamma$, such that:
\begin{equation}
	Q = \mathbb{E}[Q | \gamma, v]
\end{equation}
This indicates that the quality prediction model relies solely on the causal features $\mathbf{f}$, with $\gamma$ representing the parameter of $\phi$ in this context, excluding the influence of noise $\eta$.
The SCM, therefore, ensures that the quality assessment is both robust to noise and aligned with causal perceptual features.

Finally, on the basis of the SCM, the perceptual differences between the reference image and the distorted image are quantified via a distance metric.
Our method is predicated on the concept of perceptual cost, facilitating a quantitative analysis of the differences in perceived quality between reference and distorted images.
Specifically, we define the perceptual cost between two distributions, $P_X$ and $P_Y$, utilizing the causal transport cost, which is formally expressed as:
\begin{equation}
	COT (P_X, P_Y)= \inf _{g \in G(P_X, P_Y)}
	\int \Gamma(x,y) c(x, y) \ d \ g(x, y)
\end{equation}
where $G(P_X, P_Y)$ represents the joint distribution of the reference and distorted images, $c(x, y)$ is the l2 norm of the difference between the distributions, and $\Gamma(x,y)$ denotes the causal confounder dictionary between the two distributions.

\subsection{Connection to Existing Methods and HVS}
Existing deep learning-based methods \cite{LPIPS, DISTS, DeepWSD} leverage high-level feature representations from deep neural networks to improve perceptual alignment by statistically correlating feature distances with quality scores.
LPIPS \cite{LPIPS} measures the distance between deep network features extracted from image patches, whereas DeepWSD \cite{DeepWSD} leverages the Wasserstein distance within the deep feature space to evaluate perceptual similarity.
DISTS \cite{DISTS}, on the other hand, assesses the similarity between image structures and textures by comparing deep feature representations.
However, they still rely primarily on correlations and statistical dependencies rather than causal mechanisms, which can limit their ability to generalize effectively across diverse datasets and distortion types.
In contrast, our approach introduces a causal framework that aims to reveal the underlying mechanisms driving perceptual differences, providing a more robust and theoretically grounded approach to IQA.

HVS is known to be sensitive to specific perceptual features (e.g., contrast, texture, and structural integrity) that are often distorted by various image processing operations.
Previous studies have attempted to model HVS by designing perceptual metrics or using deep learning to automatically learn these perceptual features.
However, these models usually fail to distinguish between causal features that actually affect perceptual quality and incidental noise that may be captured by deep features.
By introducing a causal model that uses an intervention-based framework, we provide a clearer path for identifying causal features that determine perceptual quality, providing insight into the relationship between image content distortion and perceptual quality in a way that is more consistent with the mechanisms of HVS.

\begin{table*}[h]
	\renewcommand\arraystretch{1.1}
	\setlength\tabcolsep{2pt}
	\centering
	\caption{
		Results of the performance comparison conducted on benchmark datasets.
	}
	\label{tab:result}
	\resizebox{\textwidth}{!}{
		\begin{tabular}{p{2.5cm}<{\centering}p{0cm} cccccccccccccccccc}
			\toprule[1.5pt]
			\arrayrulecolor{gray} \toprule \arrayrulecolor{black}
			\multicolumn{1}{c}{\multirow{2}{*}{\makecell{Method}}}
			&& \multicolumn{2}{c}{LIVE \cite{LIVE}}
			&& \multicolumn{2}{c}{CSIQ \cite{CSIQ}}
			&& \multicolumn{2}{c}{TID2008 \cite{TID2008}}
			&& \multicolumn{2}{c}{TID2013 \cite{TID2013}}
			&& \multicolumn{2}{c}{KADID \cite{KADID}}
			&& \multicolumn{2}{c}{PIPAL \cite{PIPAL}}& \\
			\cmidrule{3-4} \cmidrule{6-7} \cmidrule{9-10} \cmidrule{12-13} \cmidrule{15-16} \cmidrule{18-19}
			&& PLCC & SRCC && PLCC & SRCC && PLCC & SRCC && PLCC & SRCC && PLCC & SRCC && PLCC & SRCC &\\
			\midrule
			PSNR
			&& 0.781 & 0.801 && 0.792 & 0.807 && 0.507 & 0.525 && 0.664 & 0.687 && 0.670 & 0.676 && 0.398 & 0.392 &\\ \rowcolor[HTML]{ececec}
			SSIM \cite{SSIM}
			&& 0.847 & 0.851 && 0.810 & 0.833 && 0.625 & 0.624 && 0.665 & 0.627 && 0.610 & 0.619 && 0.489 & 0.486 &\\
			MS-SSIM \cite{MS_SSIM}
			&& 0.886 & 0.903 && 0.875 & 0.879 && 0.842 & 0.854 && 0.831 & 0.786 && 0.824 & 0.826 && 0.571 & 0.545 &\\
			\rowcolor[HTML]{ececec}
			VIF \cite{VIF}
			&& \lred 0.949 & \lred 0.953 && 0.899 & 0.899 && 0.798 & 0.749 && 0.771 & 0.677 && 0.685 & 0.679 && 0.572 & 0.545 &\\
			MAD \cite{CSIQ}
			&& 0.904 & 0.907 && 0.922 & 0.922 && 0.681 & 0.708 && 0.737 & 0.743 && 0.717 & 0.726 && 0.614 & 0.591 &\\ \rowcolor[HTML]{ececec}
			FSIM \cite{FSIM}
			&& 0.910 & 0.920 && 0.902 & 0.915 && 0.875 & 0.884 && 0.876 & 0.851 && 0.852 & 0.854 && 0.597 & 0.573 &\\
			VSI \cite{VSI}
			&& 0.877 & 0.899 && 0.912 & 0.929 && 0.871 & 0.895 && \lblue 0.898 & \lred 0.895 && 0.877 & 0.878 && 0.548 & 0.526 &\\
			\rowcolor[HTML]{ececec}
			GMSD \cite{GMSD}
			&& 0.909 & 0.910 && \lblue 0.938 & \lblue 0.939 && 0.878 & 0.891 && 0.858 & 0.804 && 0.847 & 0.847 && 0.614 & 0.569 &\\
			NLPD \cite{NLPD}
			&& 0.882 & 0.889 && 0.913 & 0.926 && 0.866 & 0.877 && 0.832 & 0.799 && 0.809 & 0.812 && 0.489 & 0.464 &\\
			\rowcolor[HTML]{ececec}
			\midrule
			PieAPP \cite{PieAPP}
			&& 0.866 & 0.865 && 0.864 & 0.883 && 0.752 & 0.797 && 0.809 & 0.844 && 0.857 & 0.865 && \lgreen 0.702 & \lgreen 0.701 &\\
			LPIPS \cite{LPIPS}
			&& 0.866 & 0.863 && 0.891 & 0.895 && 0.722 & 0.718 && 0.713 & 0.713 && 0.838 & 0.837 && 0.584 & 0.584 &\\
			\rowcolor[HTML]{ececec}
			DISTS \cite{DISTS}
			&& 0.924 & 0.925 && 0.919 & 0.920 && 0.829 & 0.814 && 0.854 & 0.830 && 0.886* & 0.886* && 0.645 & 0.627 &\\
			DeepWSD \cite{DeepWSD}
			&& 0.904 & 0.925 && \lgreen 0.941 & \lgreen 0.950 && \lblue 0.900 & \lred 0.904 && 0.894 & 0.874 && 0.887 & 0.888 && 0.503 & 0.500 &\\
			\rowcolor[HTML]{ececec}
			TOPIQ-FR \cite{TOPIQ}
			&& 0.882 & 0.887 && 0.894 & 0.894 && 0.873 & 0.872 && 0.854 & 0.820 && \lblue 0.896 & \lblue 0.895 && \lred 0.837* & \lred 0.809* &\\
			\midrule
			Our-VGG
			&& \lgreen 0.929 & \lgreen 0.932 && \lred 0.949 & \lred 0.952 && \lred 0.905 & \lblue 0.897 && \lred 0.909 & \lgreen 0.884 && \lgreen 0.898 & \lgreen 0.899 && 0.675 & \lblue 0.657 &\\ \rowcolor[HTML]{ececec}
			Our-ResNet
			&& 0.914 & 0.921 && 0.916 & 0.923 && 0.859 & 0.868 && 0.862 & 0.852 && 0.888 & 0.890 && 0.538 & 0.524 &\\
			Our-EffNet
			&& \lblue 0.927 & \lgreen 0.932 && 0.933 & 0.938 && \lgreen 0.903 & \lgreen 0.899 && \lgreen 0.899 & \lblue 0.879 && \lred 0.905 & \lred 0.907 && \lblue 0.694 & 0.656 &\\
			\arrayrulecolor{gray} \bottomrule \arrayrulecolor{black}
			\bottomrule[1.5pt]
	\end{tabular} }
	\begin{tablenotes}
		\item Performance comparison between the results obtained by our method and those of the state-of-the-art methods. In each column, the \textcolor{red!60!black}{best}, \textcolor{green!50!black}{second}-best and \textcolor{blue!60!black}{third}-best results are highlighted in \textcolor{red!60!black}{red}, \textcolor{green!50!black}{green} and \textcolor{blue!60!black}{blue}, respectively.
		Notably, DISTS and TOPIQ-FR were trained on the KADID \cite{KADID} and PIPAL \cite{PIPAL}, respectively, which are marked with an asterisk (*) in the table for these methods.
		PieAPP and LPIPS were individually trained on their proposed datasets \cite{PieAPP} and \cite{LPIPS}, respectively.

	\end{tablenotes}
\end{table*}

\section{Experimental Results}

\subsection{Experimental Settings}
The study employs six widely used IQA databases to evaluate the effectiveness of the proposed method.
These databases include the LIVE IQA \cite{LIVE} database, CSIQ \cite{CSIQ}, TID2008 \cite{TID2008}, TID2013 \cite{TID2013}, KADID \cite{KADID}, and PIPAL \cite{PIPAL}, each containing different types of image distortions and varying numbers of reference images.
LIVE \cite{LIVE} comprises 29 reference images and 779 distorted images across five types of distortions. It provides quality scores on a scale of [0, 100].
Containing 30 reference images and 866 distorted images, CSIQ \cite{CSIQ} covers six distortion types, with subjective quality scores normalized to the range [0, 1].
TID2008 \cite{TID2008} and TID2013 \cite{TID2013} include 25 reference images. TID2008 has 1,700 distorted images with 17 types of distortions, whereas TID2013 expands to 3,000 distorted images and 24 distortion types. Both databases use a quality score range of [0, 9].
KADID \cite{KADID} provides a large-scale collection of 81 reference images and 10,125 distorted images, spanning 25 different distortion types. The quality scores fall within the interval [1, 5].
As one of the most extensive datasets, PIPAL \cite{PIPAL} includes 250 reference images and 25,850 distorted images across 40 distortion types. The subjective scores range from 868--1857.
Notably, we used only 24200 images from the training and validation sets since the labels of the test set were not available.
Given the diversity in score ranges across these databases, a linear normalization process is applied to standardize the MOS values, ensuring consistency in the evaluation process.
This step ensures that a fair comparison of the quality of predictions can be made across different datasets.

In this study, we evaluate a variety of representative and state-of-the-art image quality assessment techniques, including both feature-based and deep learning approaches, 
including the
PSNR,
SSIM \cite{SSIM},
MS-SSIM \cite{MS_SSIM},
VIF \cite{VIF},
MAD \cite{CSIQ},
FSIM \cite{FSIM},
VSI \cite{VSI},
GMSD \cite{GMSD},
NLPD \cite{NLPD},
PieAPP \cite{PieAPP},
LPIPS \cite{LPIPS},
DISTS \cite{DISTS},
DeepWSD \cite{DeepWSD},
and TOPIQ-FR \cite{TOPIQ}.
All the comparative methods were re-evaluated in our study. Specifically, for the deep learning-based approaches, we utilized the weight files provided by the authors of these methods to assess their performance through cross-dataset predictions.
The training datasets used for these weights are indicated in the table.
For the deep domain methods that do not require training, we followed the same protocol as traditional methods and conducted tests directly on the entire dataset.
To evaluate the performance of these methods, we employ the Pearson linear correlation coefficient (PLCC) and the Spearman rank correlation coefficient (SRCC) as the key metrics.
The PLCC is used to assess the accuracy of the predictions, whereas the SRCC evaluates the monotonic relationship by examining the rank correlation between the predicted quality scores and the corresponding MOS values.

Our proposed method is training-free and can be seamlessly applied to various deep learning models for image quality assessment.
Specifically, we utilize VGG \cite{VGG}, ResNet \cite{ResNet} and EfficientNet \cite{EffNet} as backbone networks to extract deep feature representations.
These networks are pretrained on large-scale datasets; thus, the features can be utilized to assess image quality without additional training.

\begin{figure*}[t]
	\centering
	\captionsetup[subfloat]{font=scriptsize}
	\renewcommand{\thesubfigure}{\arabic{subfigure}}
	\subfloat[SSIM / LIVE]{\includegraphics[width=0.165\linewidth]{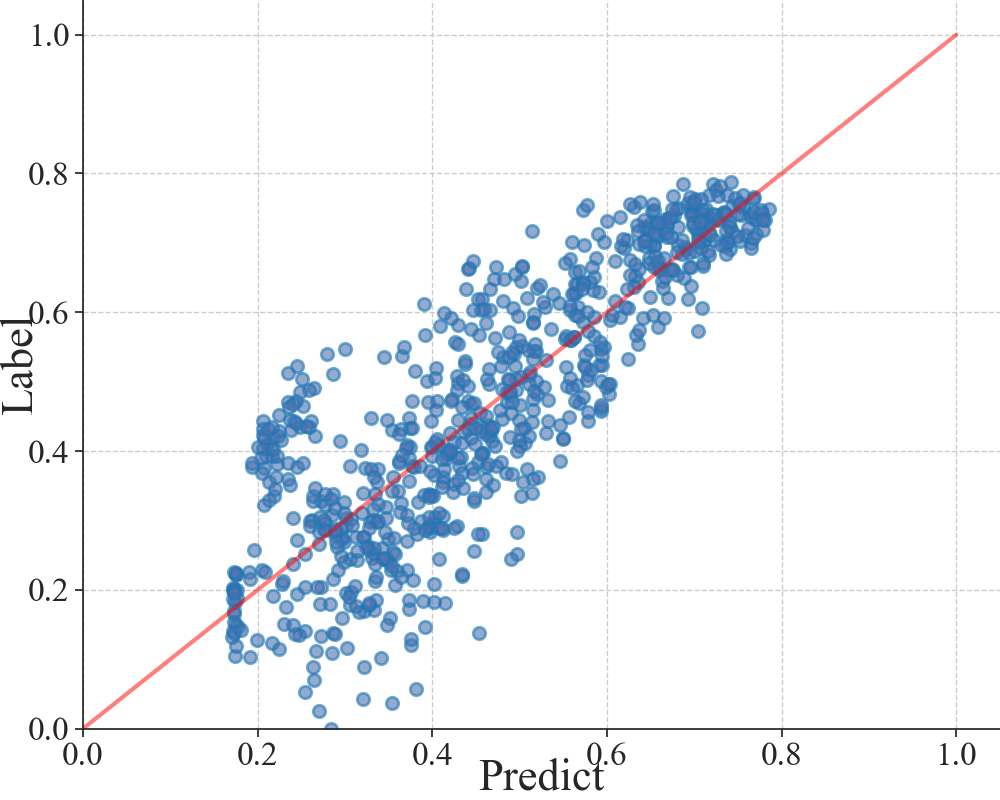}}
	\subfloat[SSIM / CSIQ]{\includegraphics[width=0.165\linewidth]{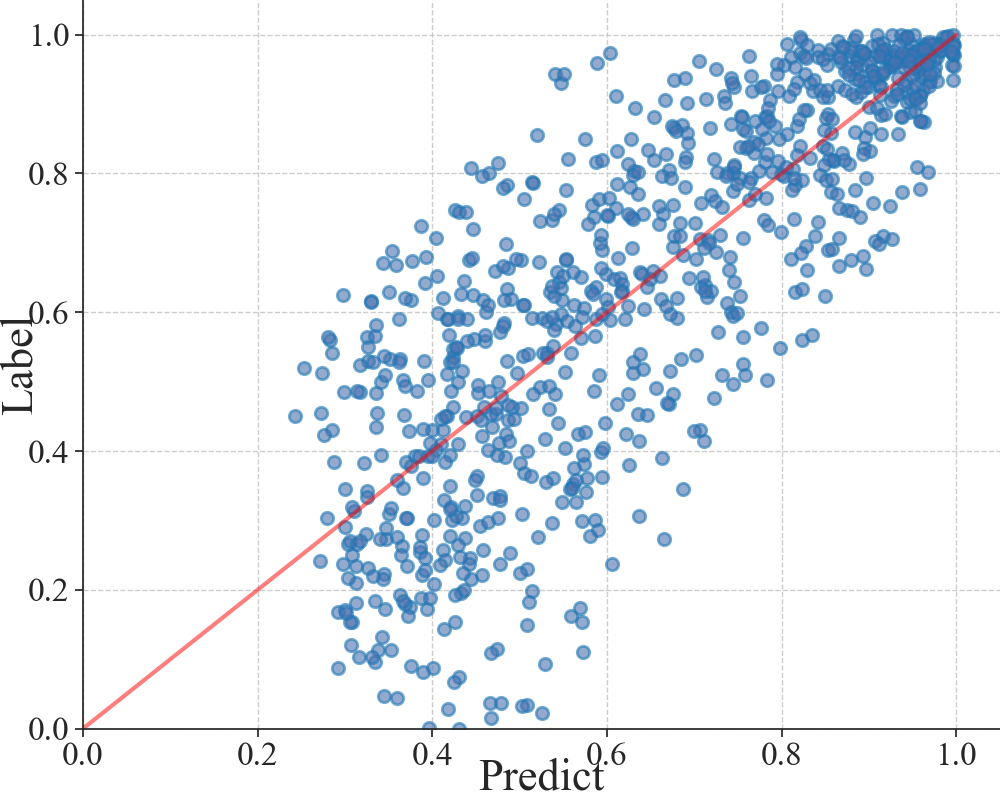}}
	\subfloat[SSIM / TID2008]{\includegraphics[width=0.165\linewidth]{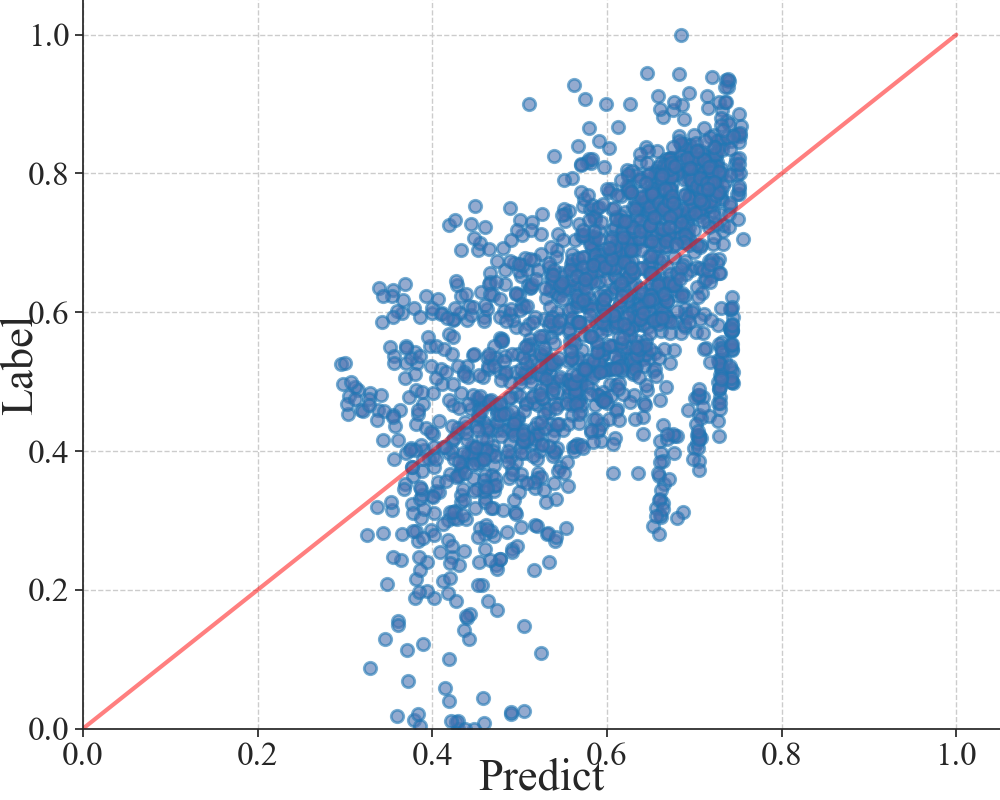}}
	\subfloat[SSIM / TID2013]{\includegraphics[width=0.165\linewidth]{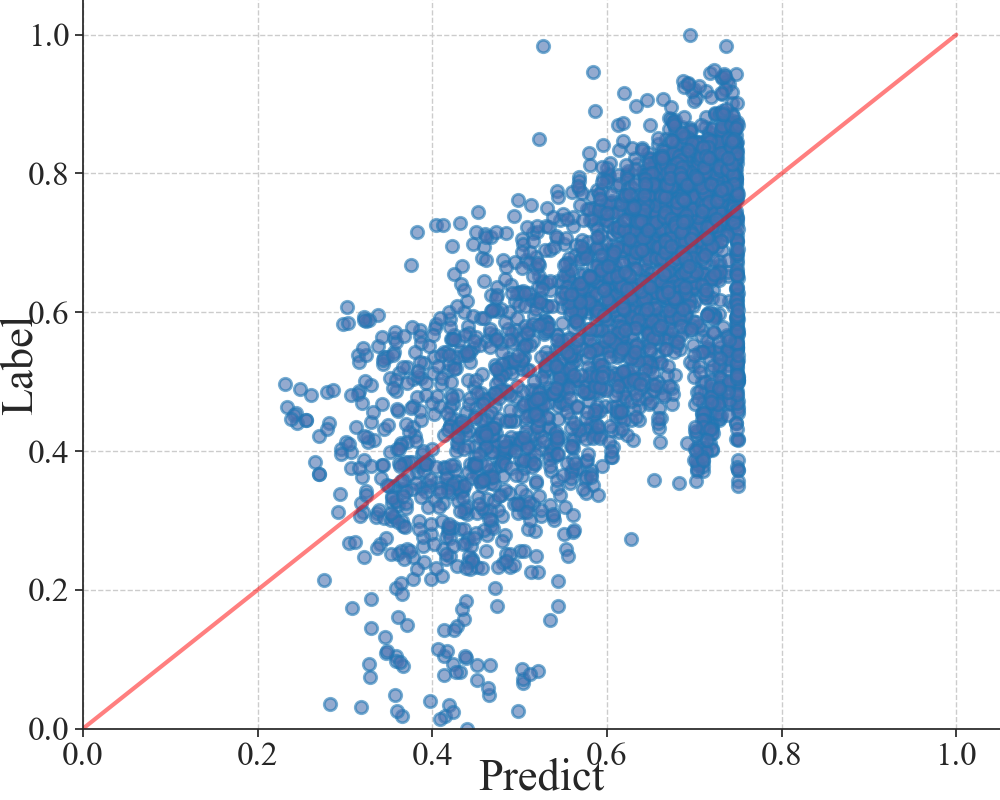}}
	\subfloat[SSIM / KADID]{\includegraphics[width=0.165\linewidth]{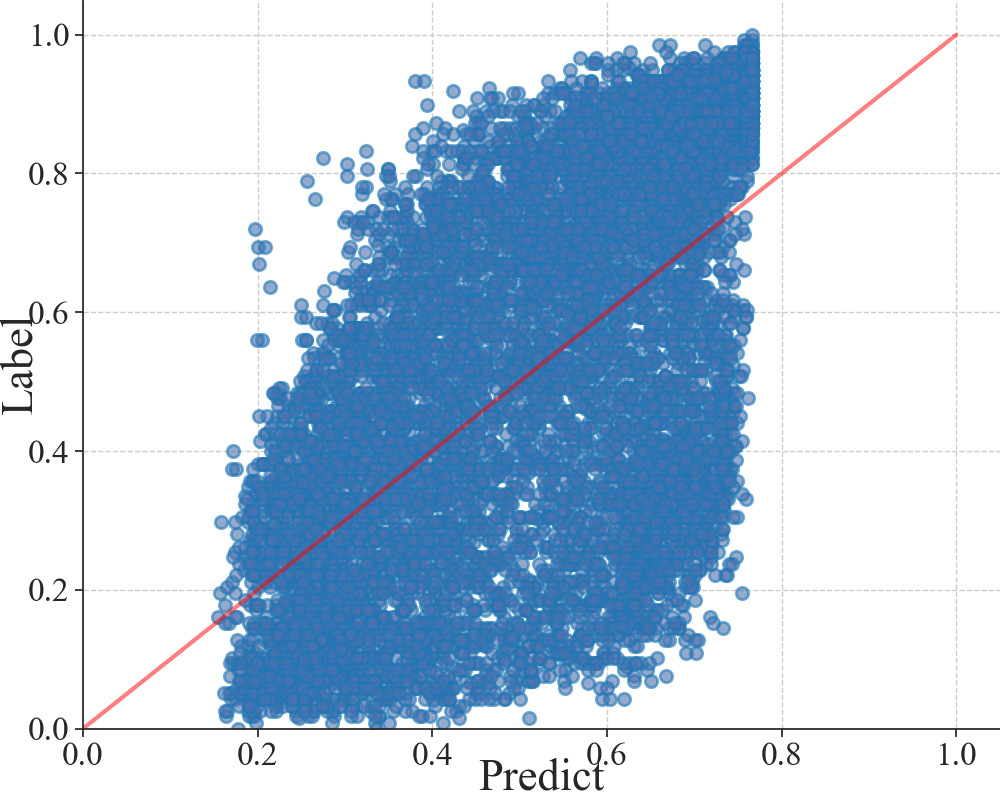}}
	\subfloat[SSIM / PIPAL]{\includegraphics[width=0.165\linewidth]{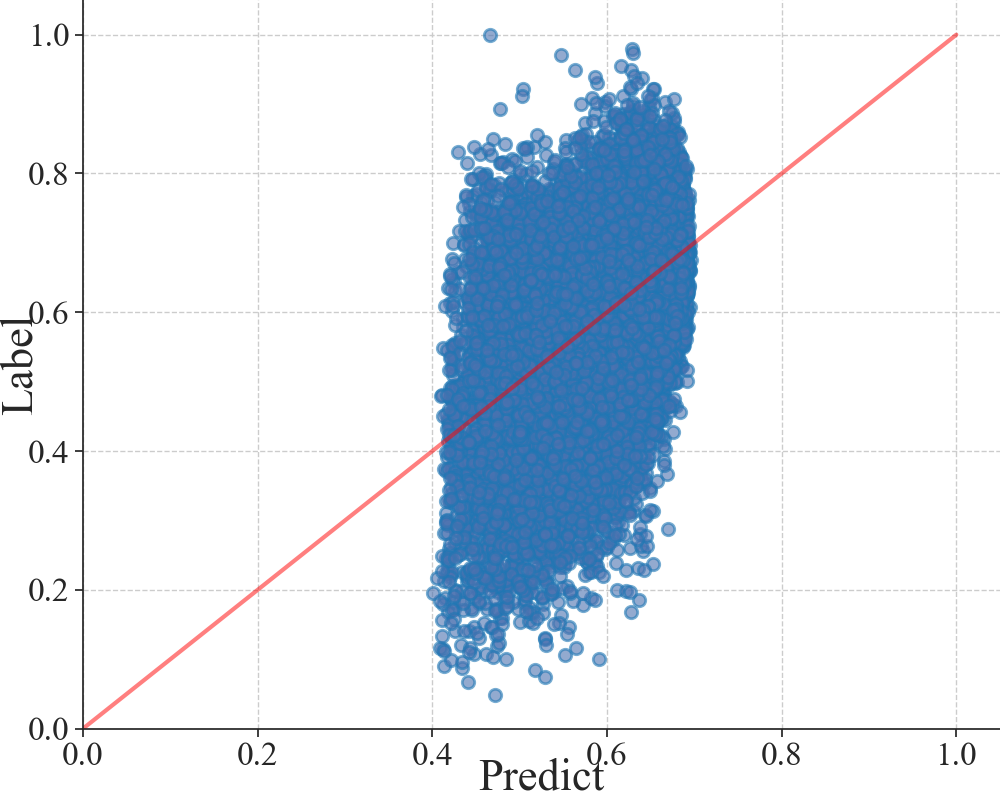}}
	\vspace{0.5em} \\
	\subfloat[DISTS / LIVE]{\includegraphics[width=0.165\linewidth]{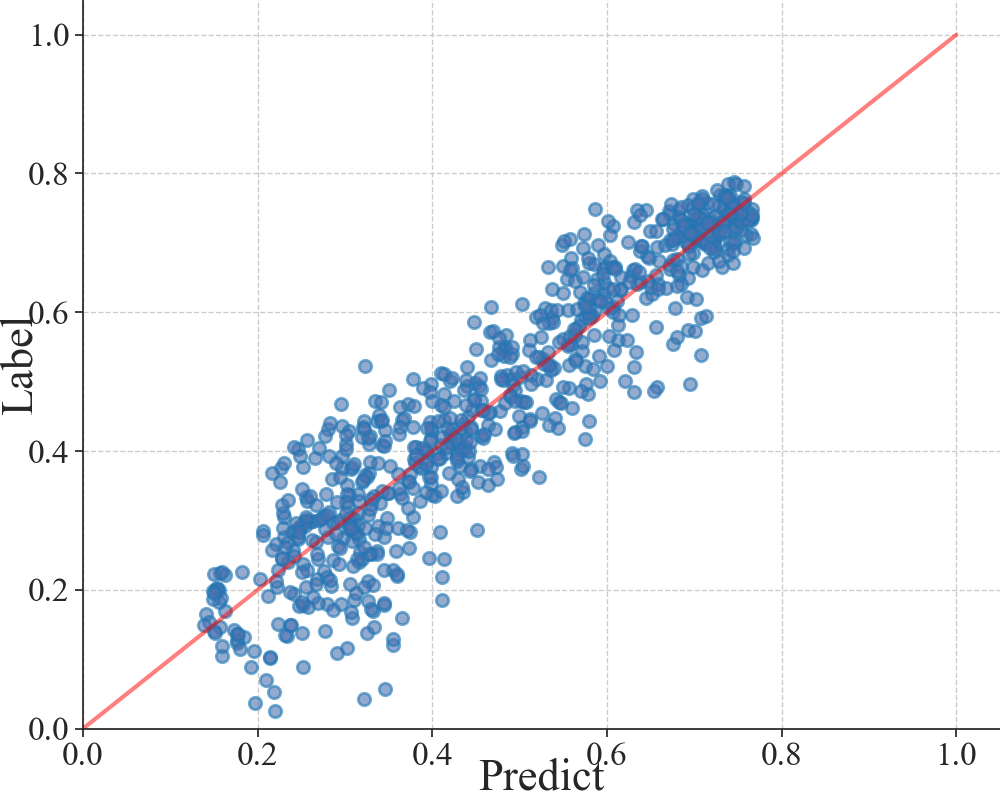}}
	\subfloat[DISTS / CSIQ]{\includegraphics[width=0.165\linewidth]{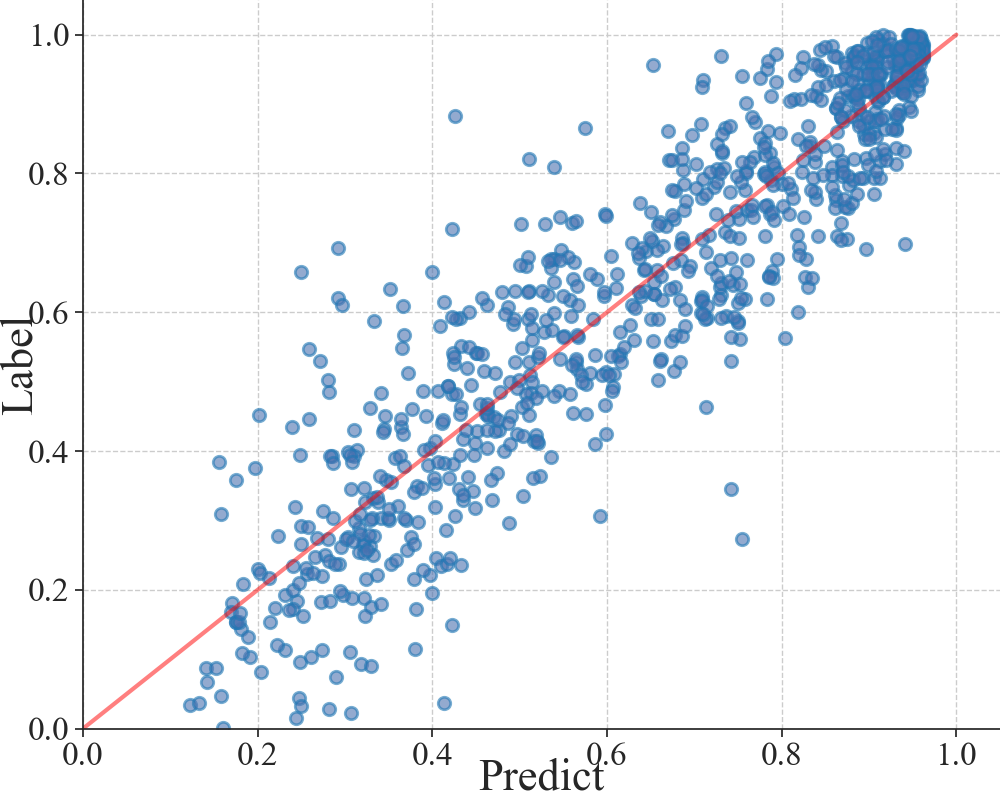}}
	\subfloat[DISTS / TID2008]{\includegraphics[width=0.165\linewidth]{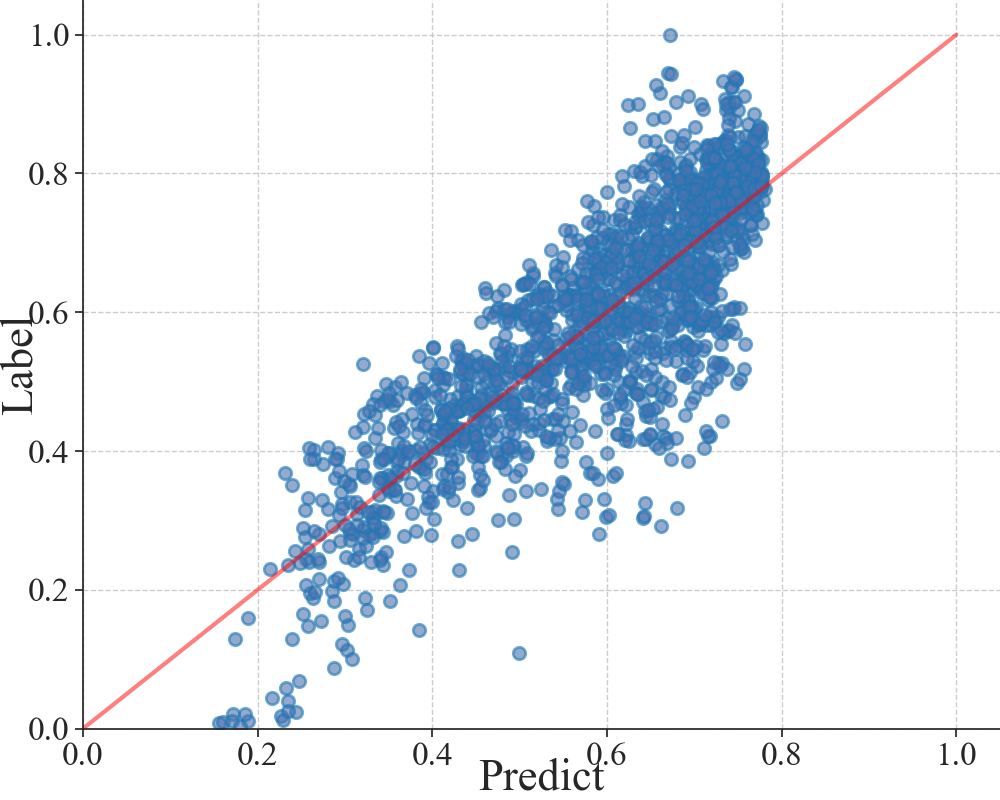}}
	\subfloat[DISTS / TID2013]{\includegraphics[width=0.165\linewidth]{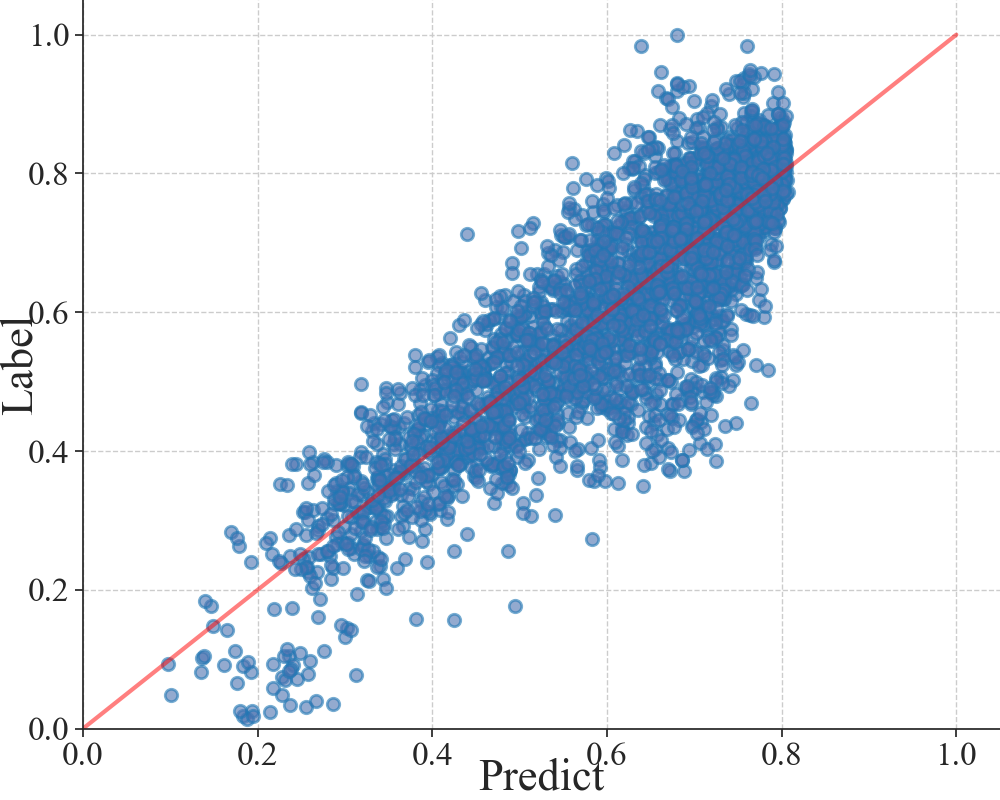}}
	\subfloat[DISTS / KADID]{\includegraphics[width=0.165\linewidth]{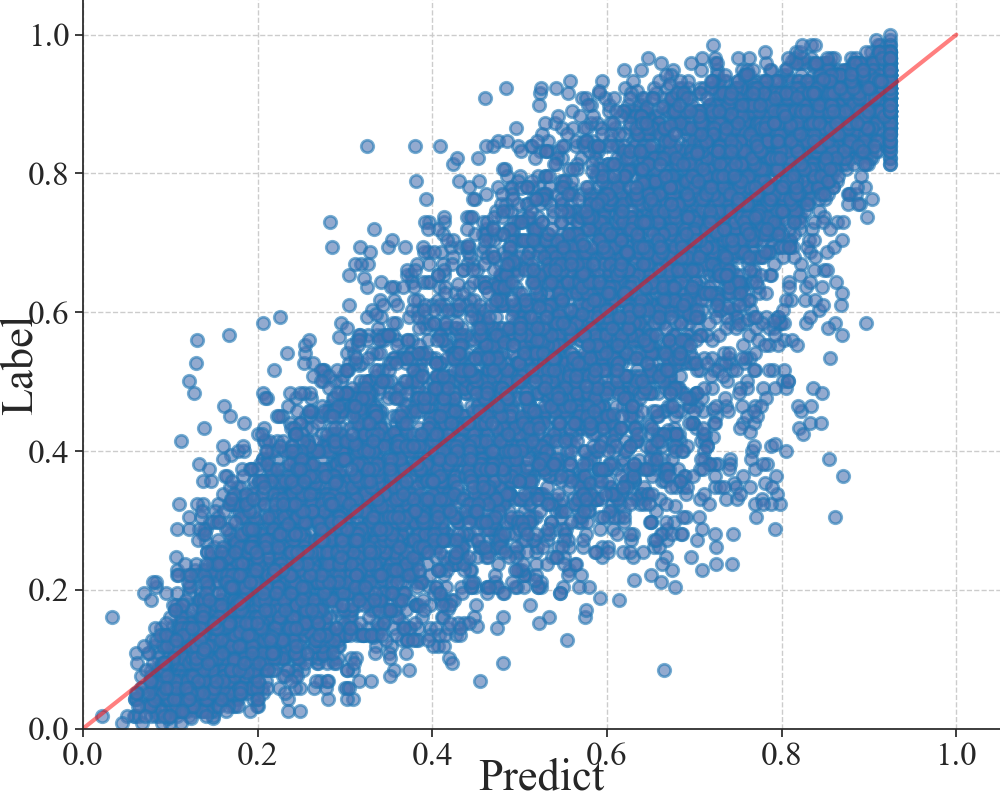}}
	\subfloat[DISTS /PIPAL ]{\includegraphics[width=0.165\linewidth]{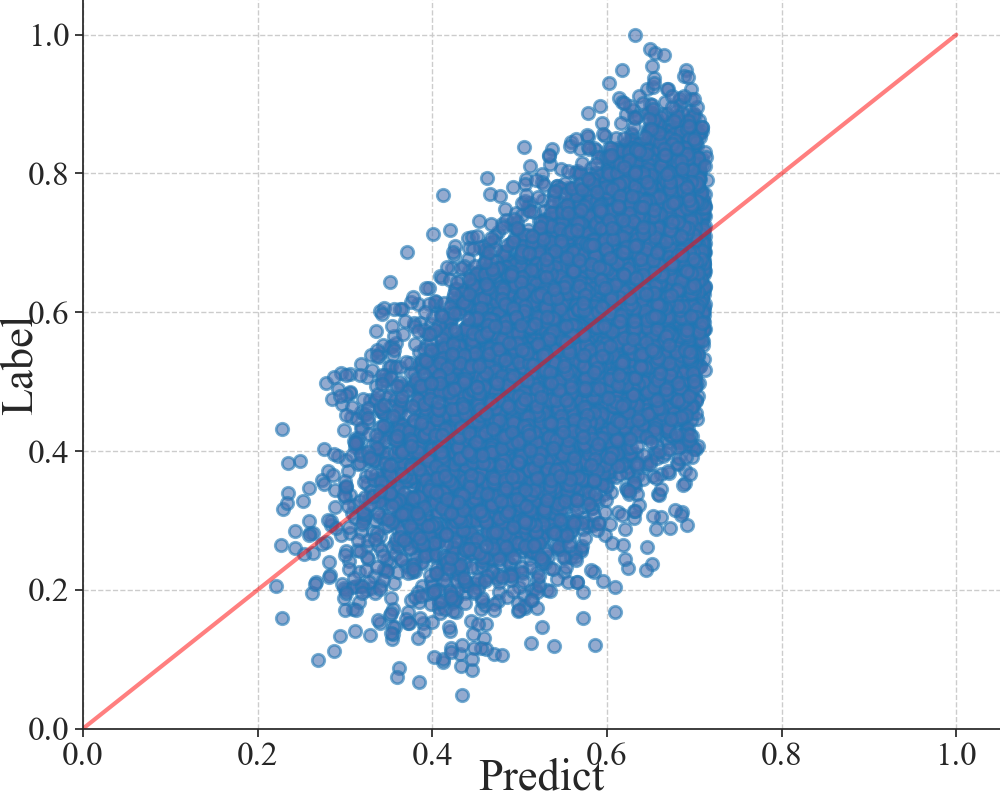}}
	\vspace{0.5em} \\
	\subfloat[DeepWSD / LIVE]{\includegraphics[width=0.165\linewidth]{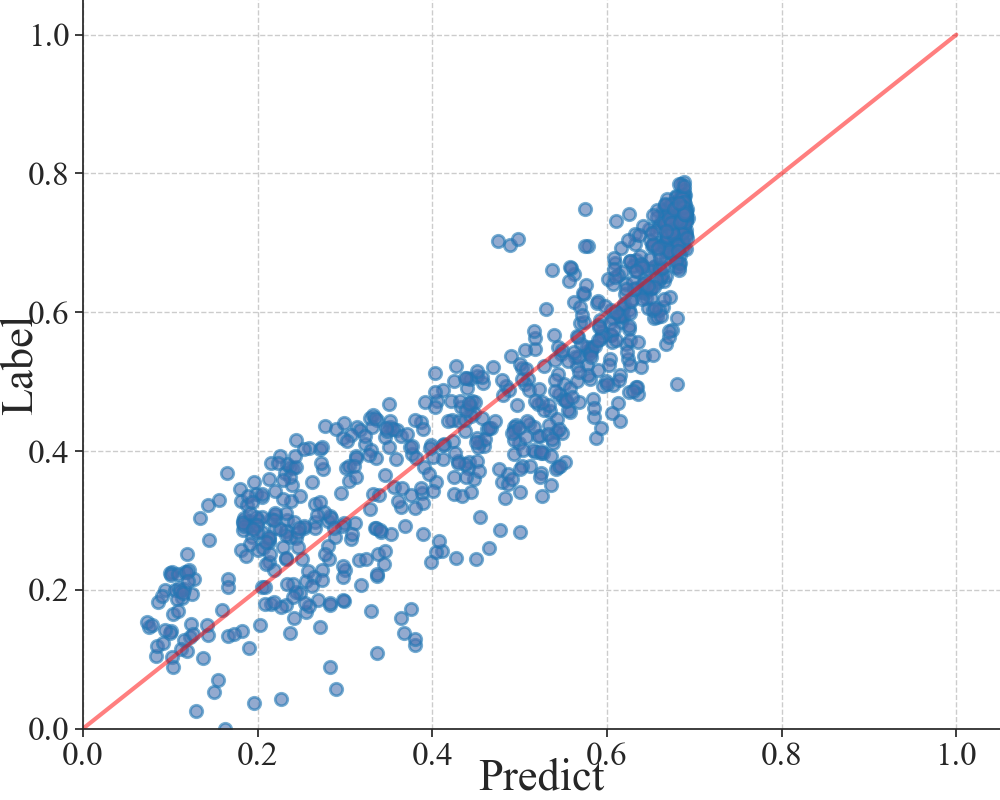}} 
	\subfloat[DeepWSD / CSIQ]{\includegraphics[width=0.165\linewidth]{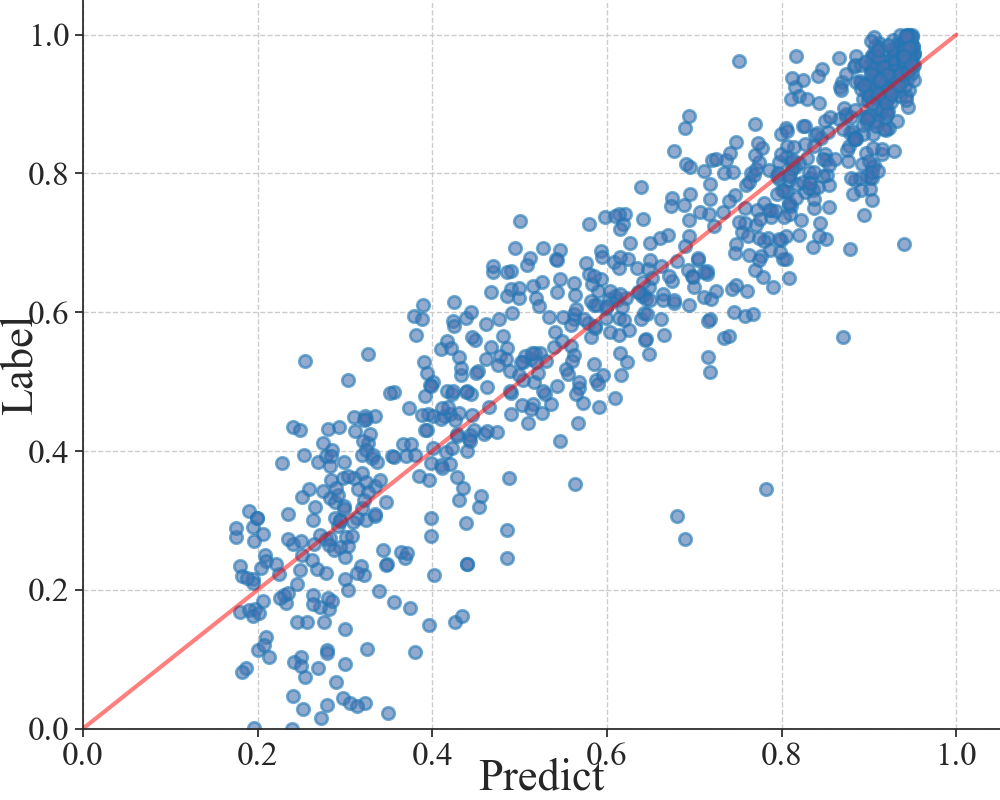}} 
	\subfloat[DeepWSD / TID2008]{\includegraphics[width=0.165\linewidth]{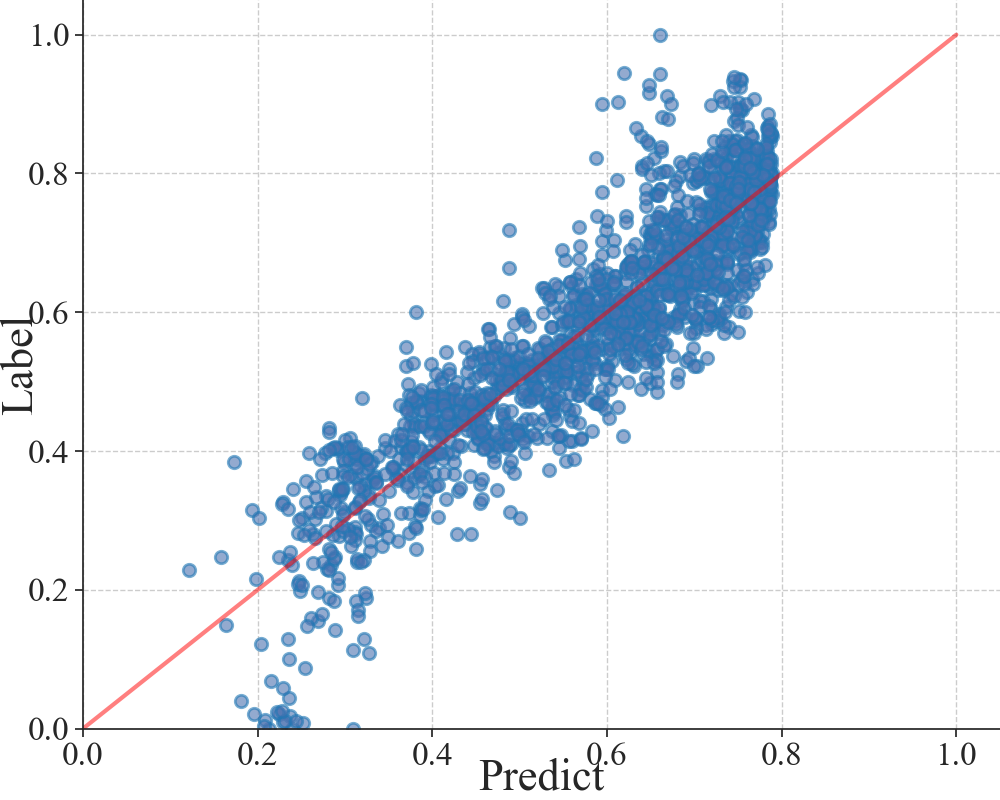}} 
	\subfloat[DeepWSD / TID2013]{\includegraphics[width=0.165\linewidth]{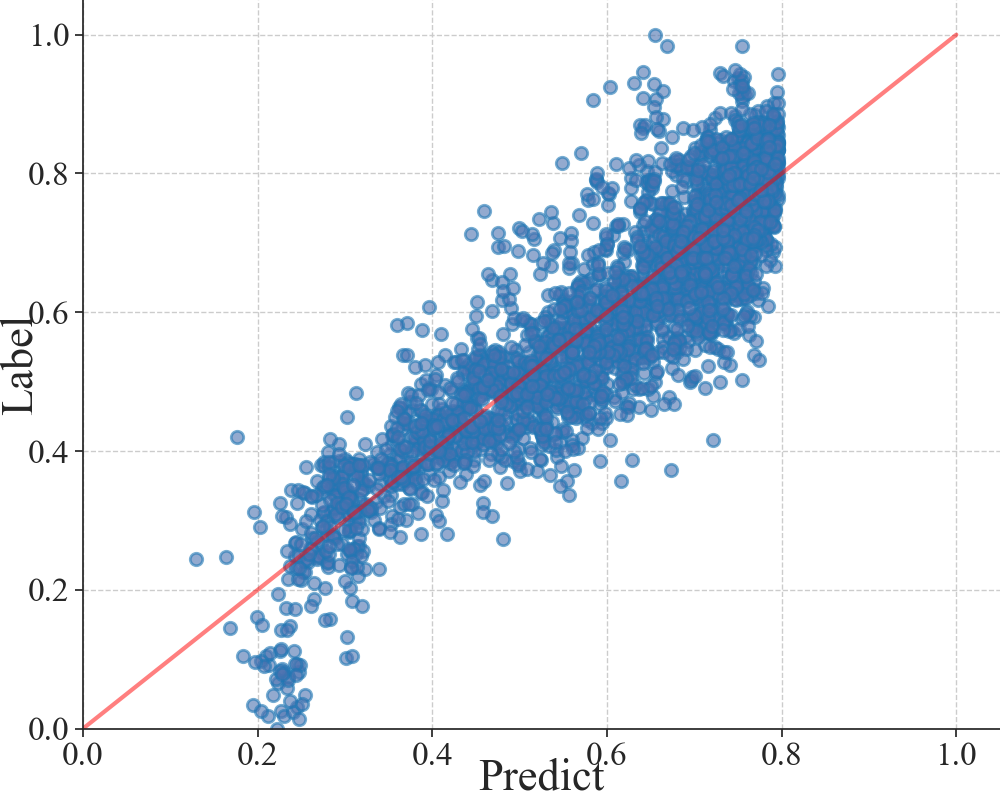}}  
	\subfloat[DeepWSD / KADID]{\includegraphics[width=0.165\linewidth]{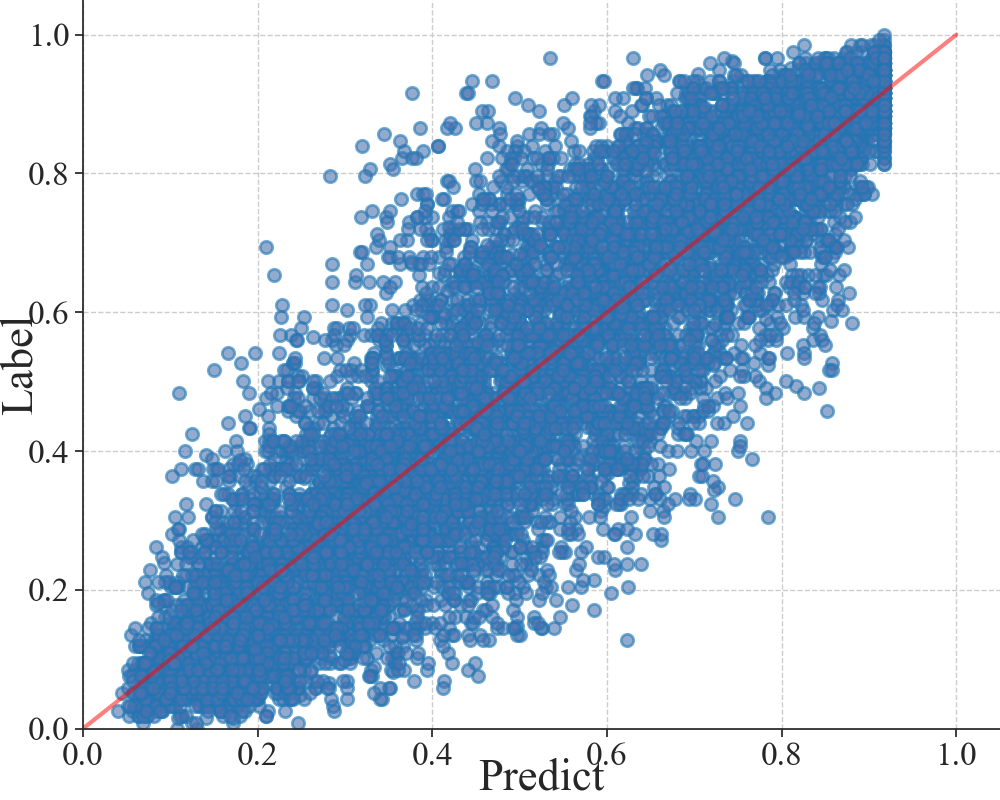}} 
	\subfloat[DeepWSD / PIPAL]{\includegraphics[width=0.165\linewidth]{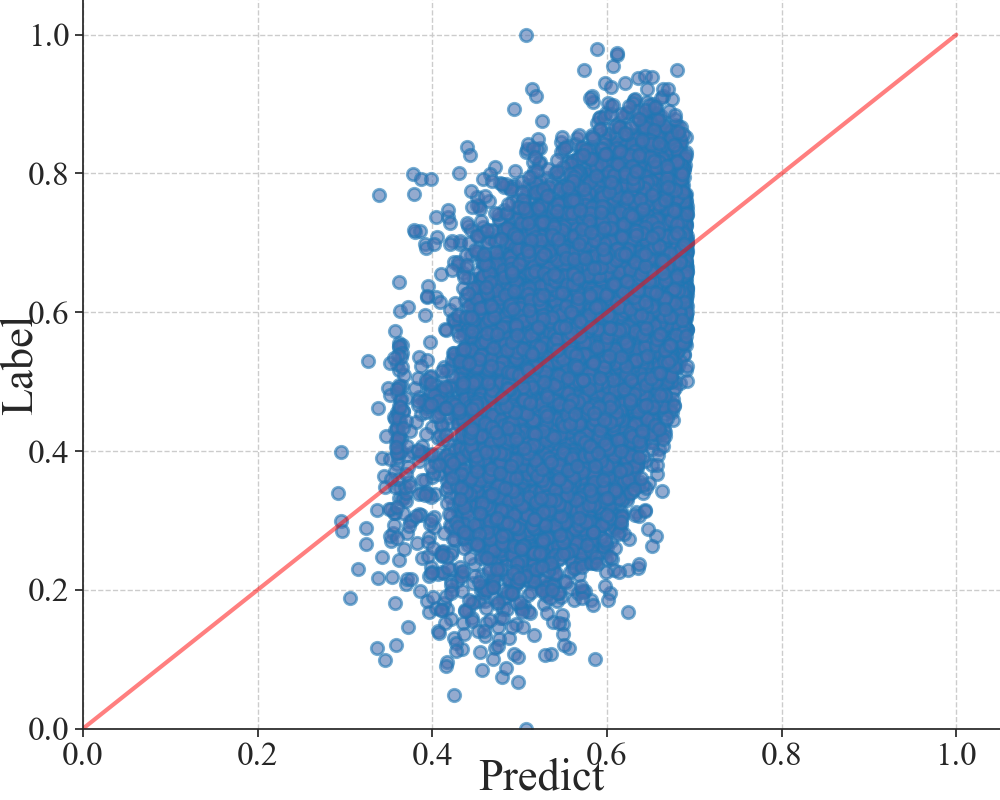}} 
	\vspace{0.5em} \\
	\subfloat[TOPIQ-FR / LIVE]{\includegraphics[width=0.165\linewidth]{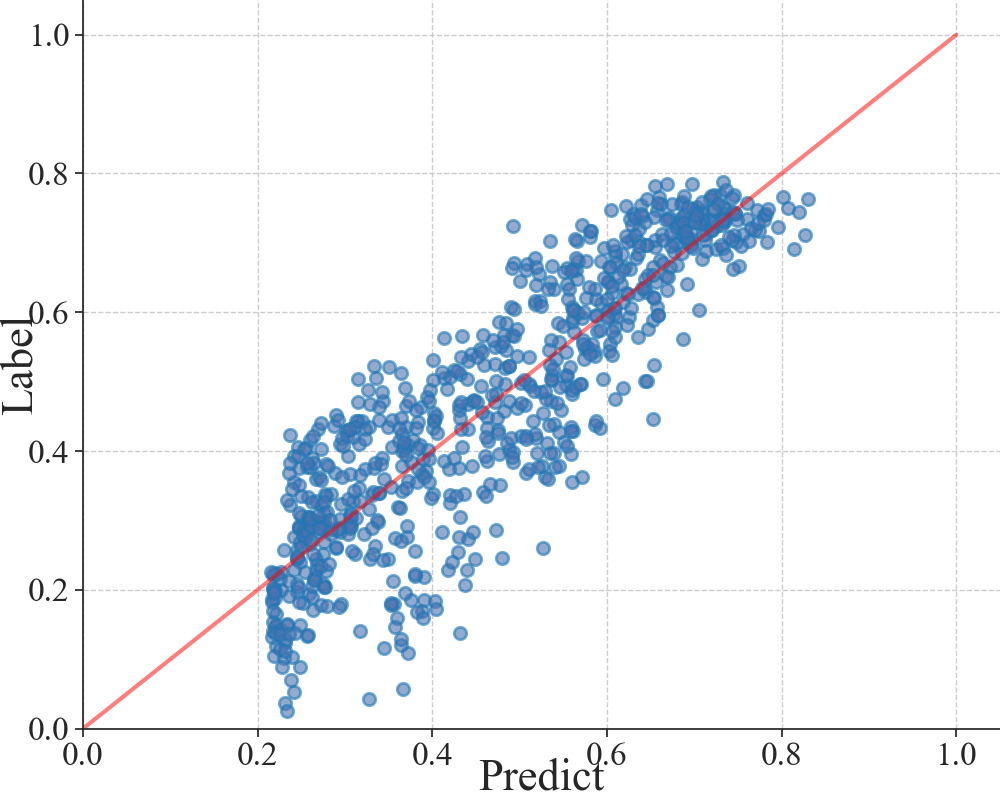}} 
	\subfloat[TOPIQ-FR / CSIQ]{\includegraphics[width=0.165\linewidth]{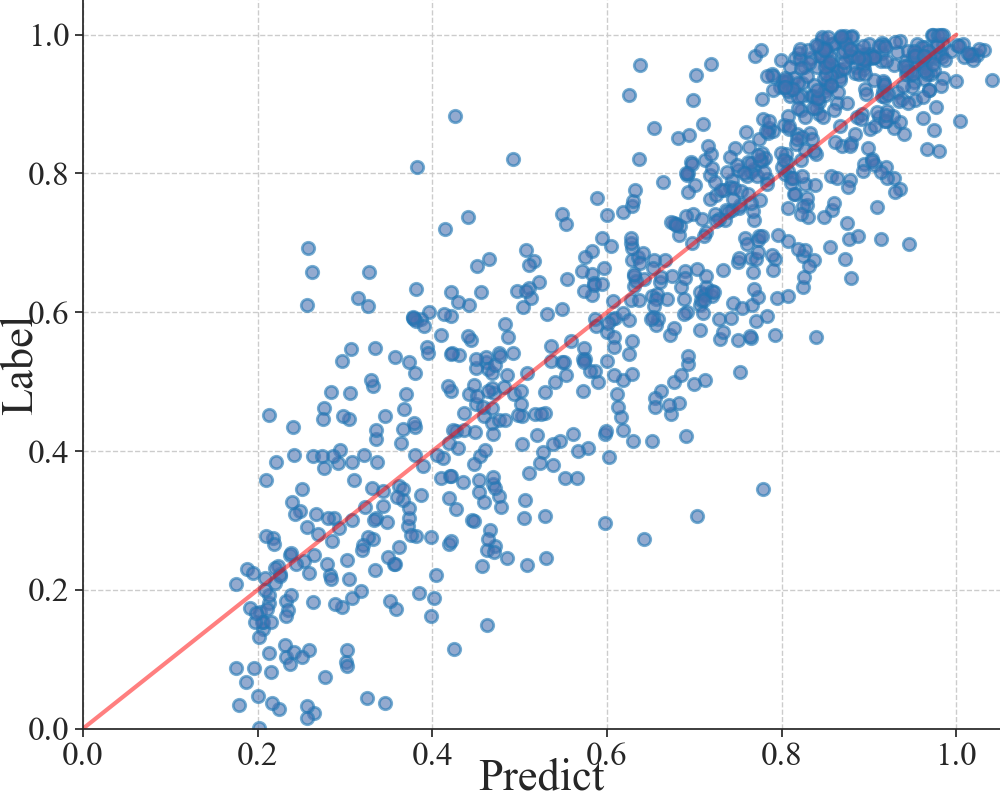}} 
	\subfloat[TOPIQ-FR / TID2008]{\includegraphics[width=0.165\linewidth]{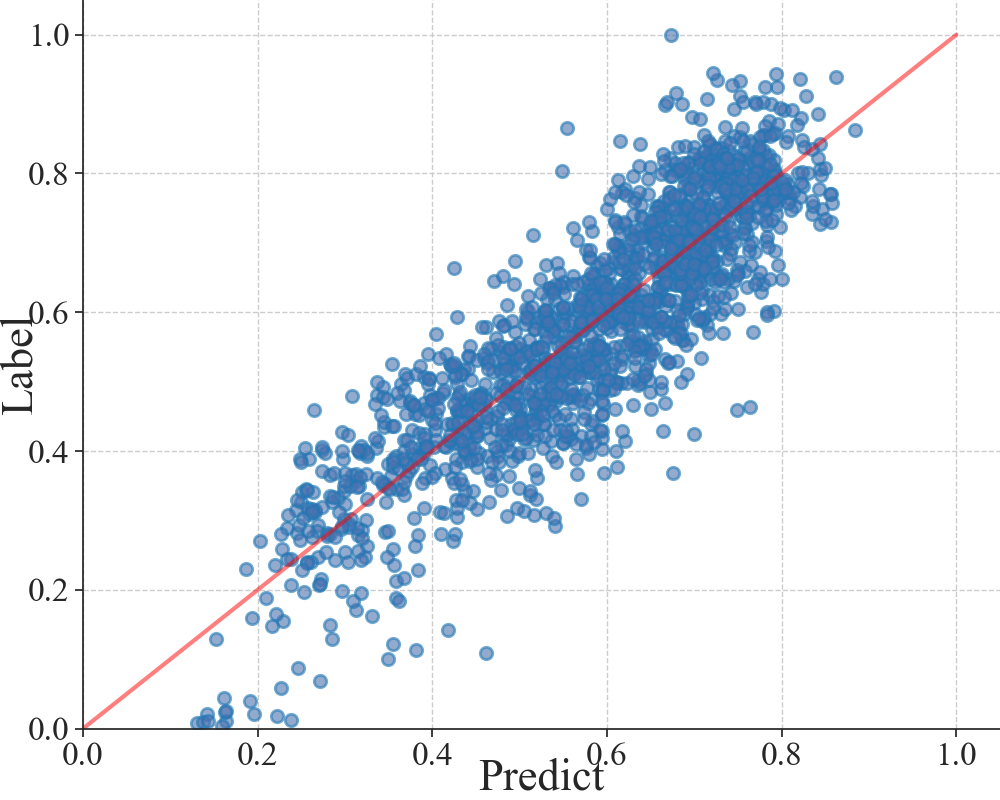}} 
	\subfloat[TOPIQ-FR / TID2013]{\includegraphics[width=0.165\linewidth]{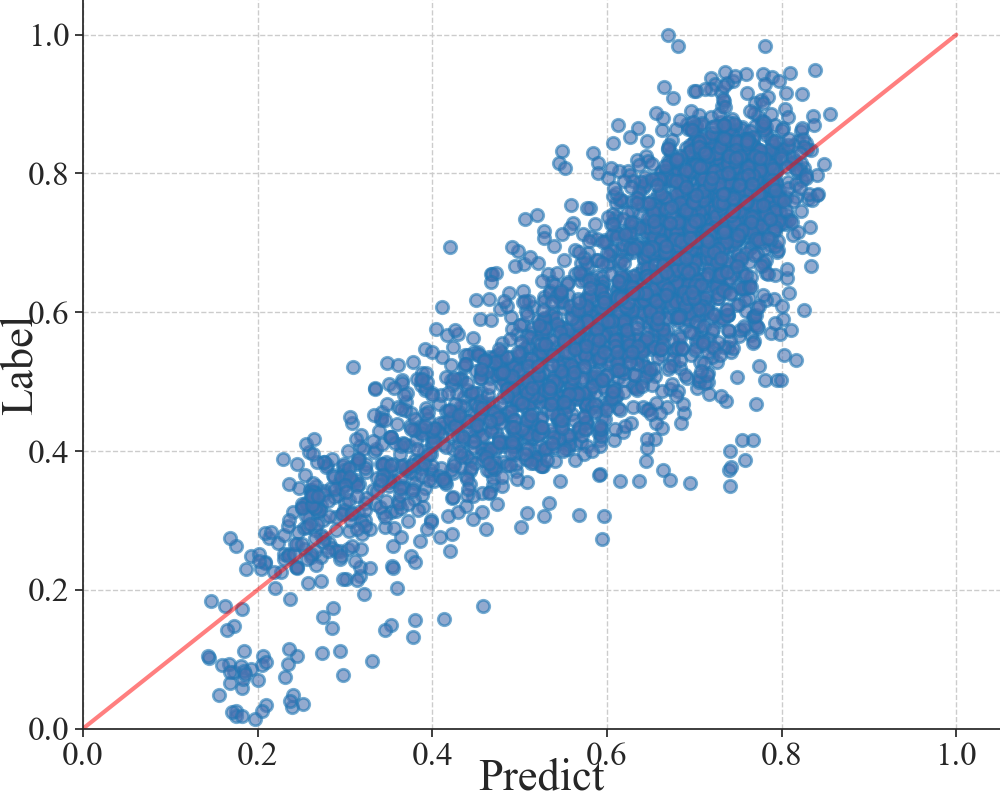}} 
	\subfloat[TOPIQ-FR / KADID]{\includegraphics[width=0.165\linewidth]{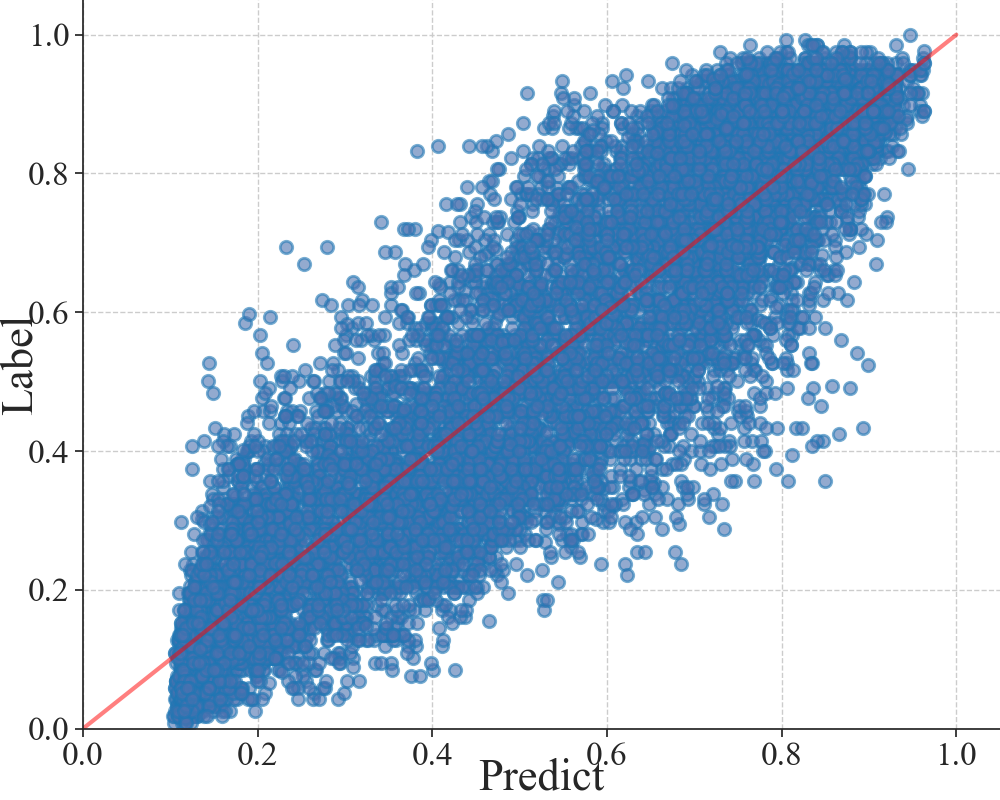}} 
	\subfloat[TOPIQ-FR / PIPAL]{\includegraphics[width=0.165\linewidth]{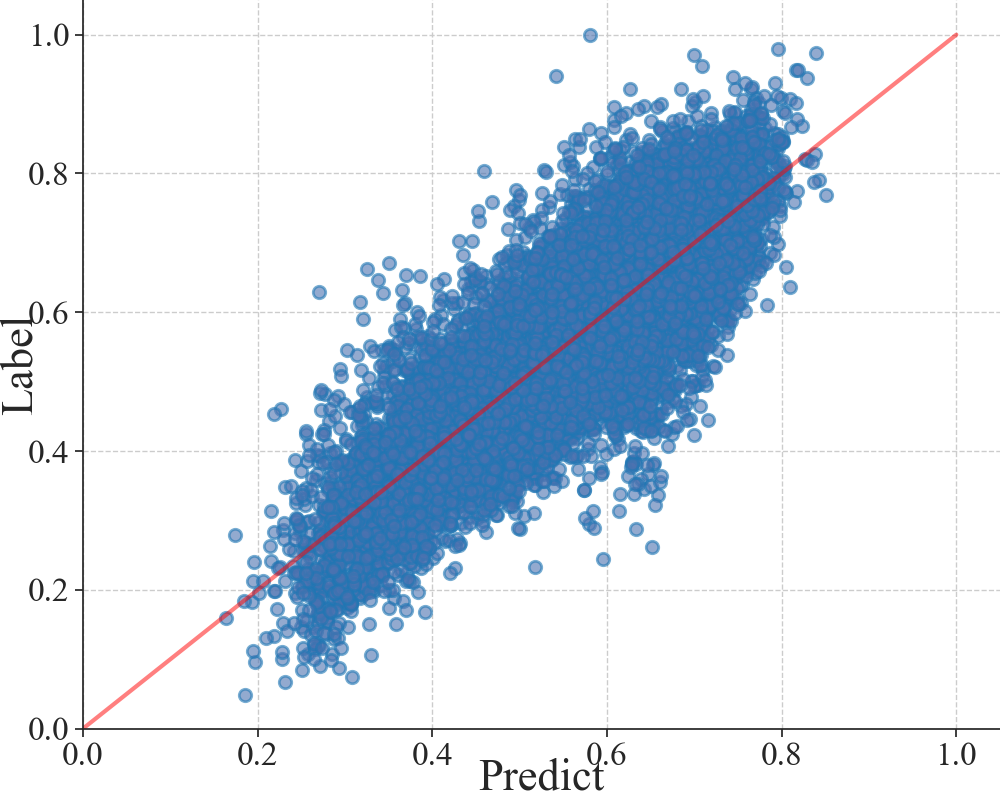}}
	\vspace{0.5em} \\
	\subfloat[Ours / LIVE]{\includegraphics[width=0.165\linewidth]{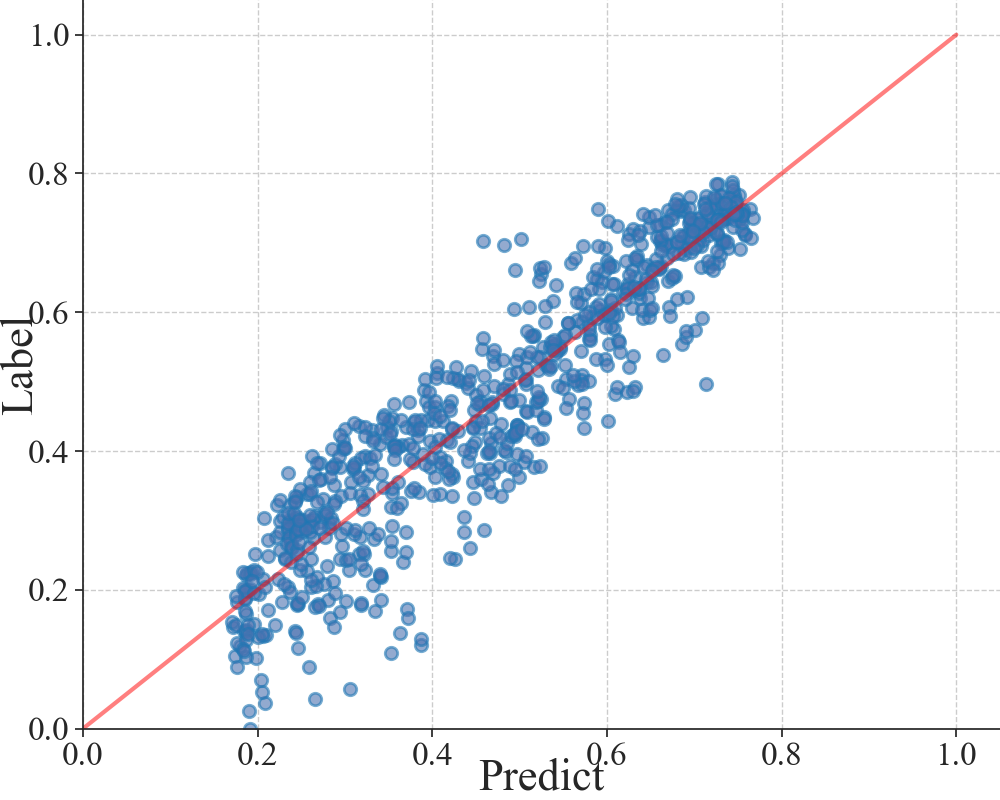}} 
	\subfloat[Ours / CSIQ]{\includegraphics[width=0.165\linewidth]{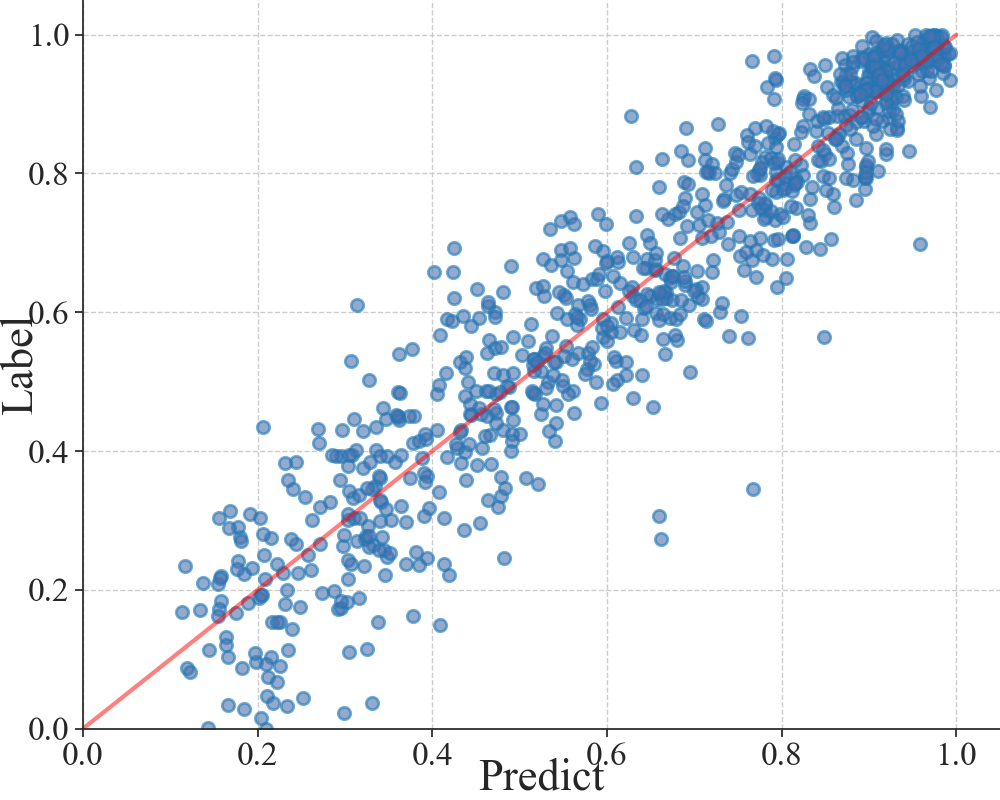}} 
	\subfloat[Ours / TID2008]{\includegraphics[width=0.165\linewidth]{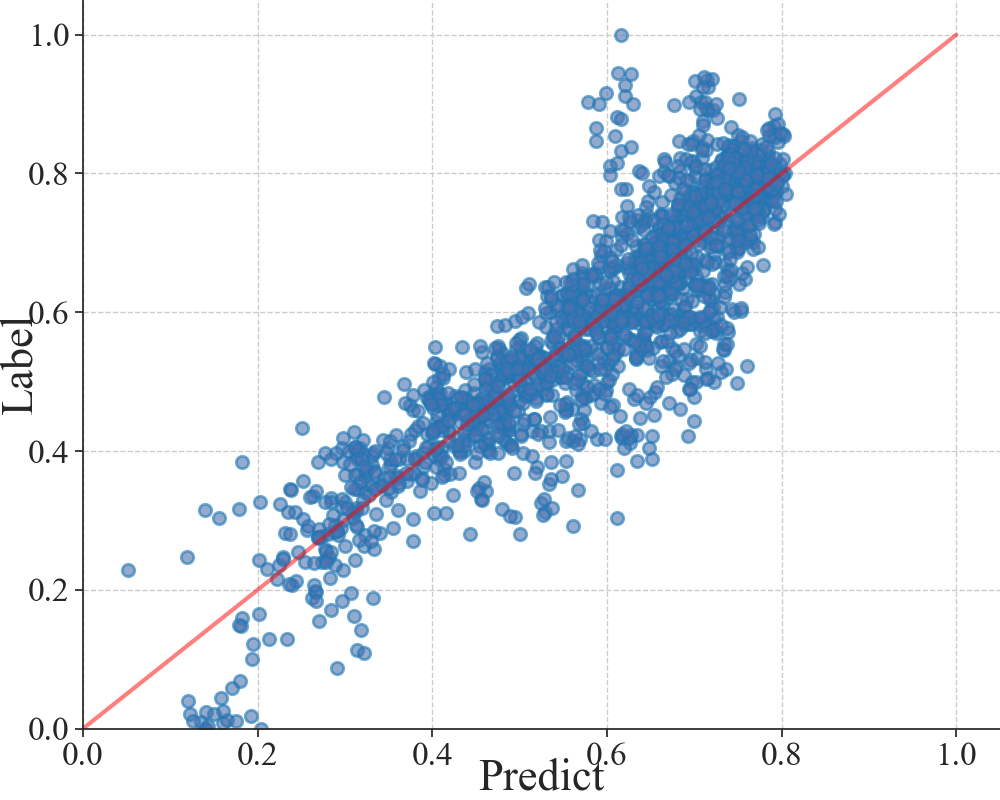}} 
	\subfloat[Ours / TID2013]{\includegraphics[width=0.165\linewidth]{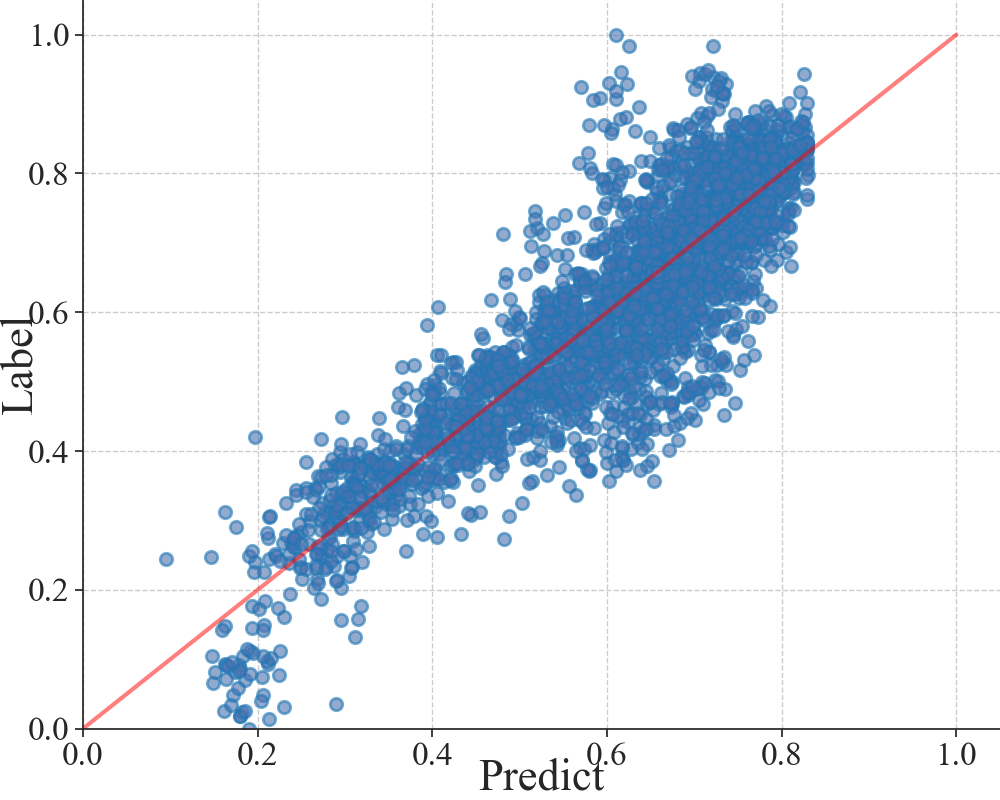}} 
	\subfloat[Ours / KADID]{\includegraphics[width=0.165\linewidth]{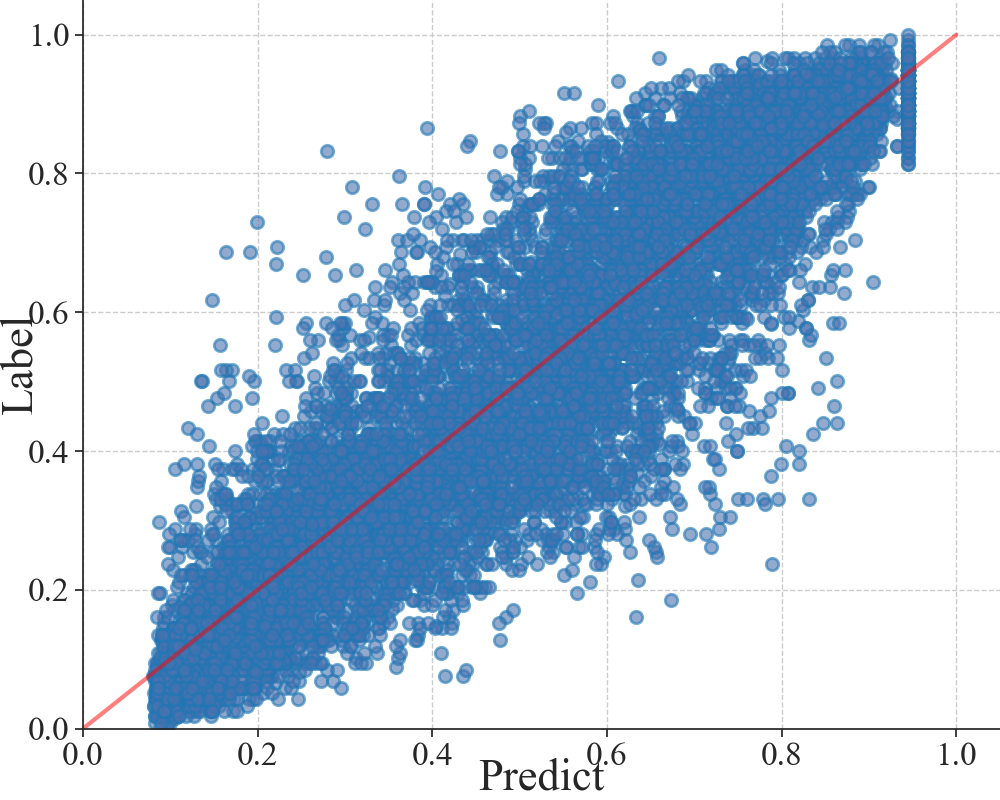}} 
	\subfloat[Ours / PIPAL]{\includegraphics[width=0.165\linewidth]{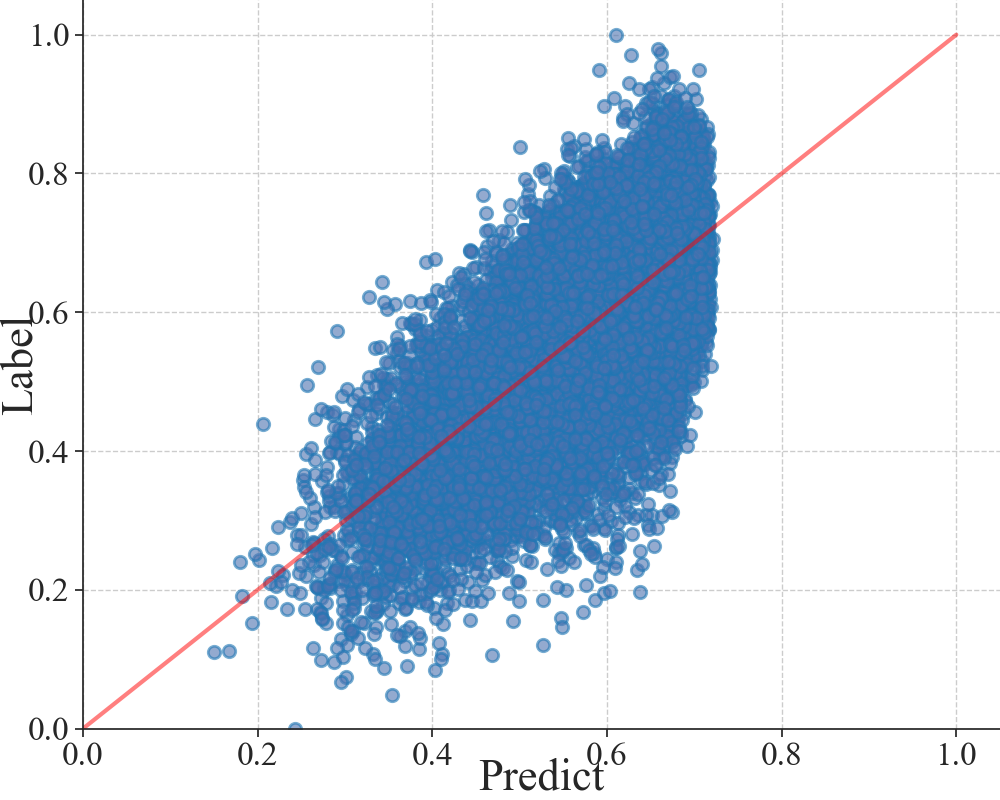}} 
	\caption{
		Scatterplot of prediction results for different methods across databases.
		We performed normalization on both labels and predictions and 4-parameter logistic function is utilized before comparison, as suggested in \cite{VQEG}, and the values of the lower-better methods are reversed before regression.
	}
	\label{fig:plot}
\end{figure*}

\begin{table}[t]
	\small
	\renewcommand\arraystretch{1}
	\setlength\tabcolsep{2pt}
	\centering
	\caption{Performance comparison results of the ablation experiments.}
	\label{tab:ablation}
	\resizebox{0.49\textwidth}{!}{
		\begin{tabular}{cp{1.5cm}<{\centering}cccccccccc}
			\toprule[1.5pt]
			\arrayrulecolor{gray} \toprule \arrayrulecolor{black}
			& Model
			&& \multicolumn{2}{c}{LIVE \cite{LIVE}}
			&& \multicolumn{2}{c}{CSIQ \cite{CSIQ}}
			&& \multicolumn{2}{c}{TID2013 \cite{TID2013}} & \\
			\cmidrule{4-5} \cmidrule{7-8} \cmidrule{10-11}
			& Parameter && PLCC  & SRCC  && PLCC   & SRCC   && PLCC   & SRCC & \\
			\midrule
			& $\phi = \phi_\theta$
			&& 0.901 & 0.915 && 0.913 & 0.916 && 0.884 & 0.867 &\\ \rowcolor[HTML]{ececec}
			& $\phi = \phi_\gamma$
			&& 0.929 & 0.932 && 0.949 & 0.952 && 0.909 & 0.884 &\\
			& $\phi = \phi_\eta$
			&& 0.843 & 0.866 && 0.803 & 0.831 && 0.786 & 0.789  &\\ \rowcolor[HTML]{ececec}
			\arrayrulecolor{gray} \bottomrule \arrayrulecolor{black}
			\bottomrule[1.5pt]
	\end{tabular} }
	\begin{tablenotes}
		\item Performance comparison results of the ablation experiments conducted on three benchmark datasets, where we use Our-VGG as the default architecture.
		The models evaluated in this study correspond to different configurations of weight assignments for $\phi$, where its values $\phi_\theta$, $\phi_\gamma$, and $\phi_\eta$ represent the original pretrained weights, the causal effect weights, and the residual component excluding the causal effect, respectively.
	\end{tablenotes}
\end{table}

\subsection{Quality Prediction Performance}
We conduct quality prediction experiments to evaluate the performance and generalizability of our model in comparison with several state-of-the-art methods. The methods we compare can be categorized into three main groups: traditional image quality assessment methods, end-to-end deep learning models, and deep feature-based similarity (or difference) metrics that do not require training.
As presented in \autoref{tab:result} and visualized in the scatter plot in \autoref{fig:plot}, our method demonstrates superior overall performance and consistently aligns closely with actual scores across datasets.
Traditional techniques, which are typically based on handcrafted features, often struggle to capture the complex, nonlinear relationships between image distortions and perceptual quality, resulting in relatively lower accuracy across diverse datasets.
End-to-end deep learning methods, while showing a marked improvement over traditional methods, exhibit notable limitations in terms of generalizability. These methods are typically trained on specific datasets, and as a result, their performance tends to degrade when evaluated on datasets that differ from the training set, struggling to maintain consistency and accuracy when applied to unseen data.
Furthermore, the deep feature-based similarity methods that do not rely on training also perform well, although they tend to be less precise than our model is. These methods capture image distortions via pretrained features, but their inability to explicitly model causal relationships results in slightly less accurate predictions, particularly when the perceptual impact of distortions varies significantly across datasets.
In contrast, our approach, which leverages causal reasoning through an SCM and does not require training, excels in maintaining consistent performance across various datasets. The results demonstrate that, by focusing on causally relevant features, our model is less sensitive to the idiosyncrasies of individual datasets and is capable of producing accurate predictions for unseen images. This robustness in generalization suggests that the causal framework used in our model provides a more reliable representation of perceptual quality, independent of dataset-specific biases.

\subsection{Ablation Analysis}
In this section, we present the abductive counterfactual experiments to evaluate the impact of different weight configurations in our model. Specifically, we analyse the results when varying the weights of $\theta$, $\gamma$, and $\eta$.
Here, $\theta$ represents the original pretrained weights, $\gamma$ corresponds to the weight associated with the causal effect, and $\eta$ denotes the remaining part of the model, which is the portion of $\theta$ excluding $\gamma$.
The ablation analysis is conducted by systematically adjusting each of these weights and observing their influence on the model's performance.
The results are summarized in \autoref{tab:ablation}, which indicate that although noncausal representations also exhibit some perceptual relevance, their lack of causal relationships limits their effectiveness.
When these noncausal representations are removed from the original feature set, the model's predictive performance improves.
This suggests that incorporating those features that capture causal effects can enhance the model's ability to make more accurate and reliable predictions in the context of image quality assessment.

\begin{figure}[t]
	\centering
	\includegraphics[width=0.5\textwidth]{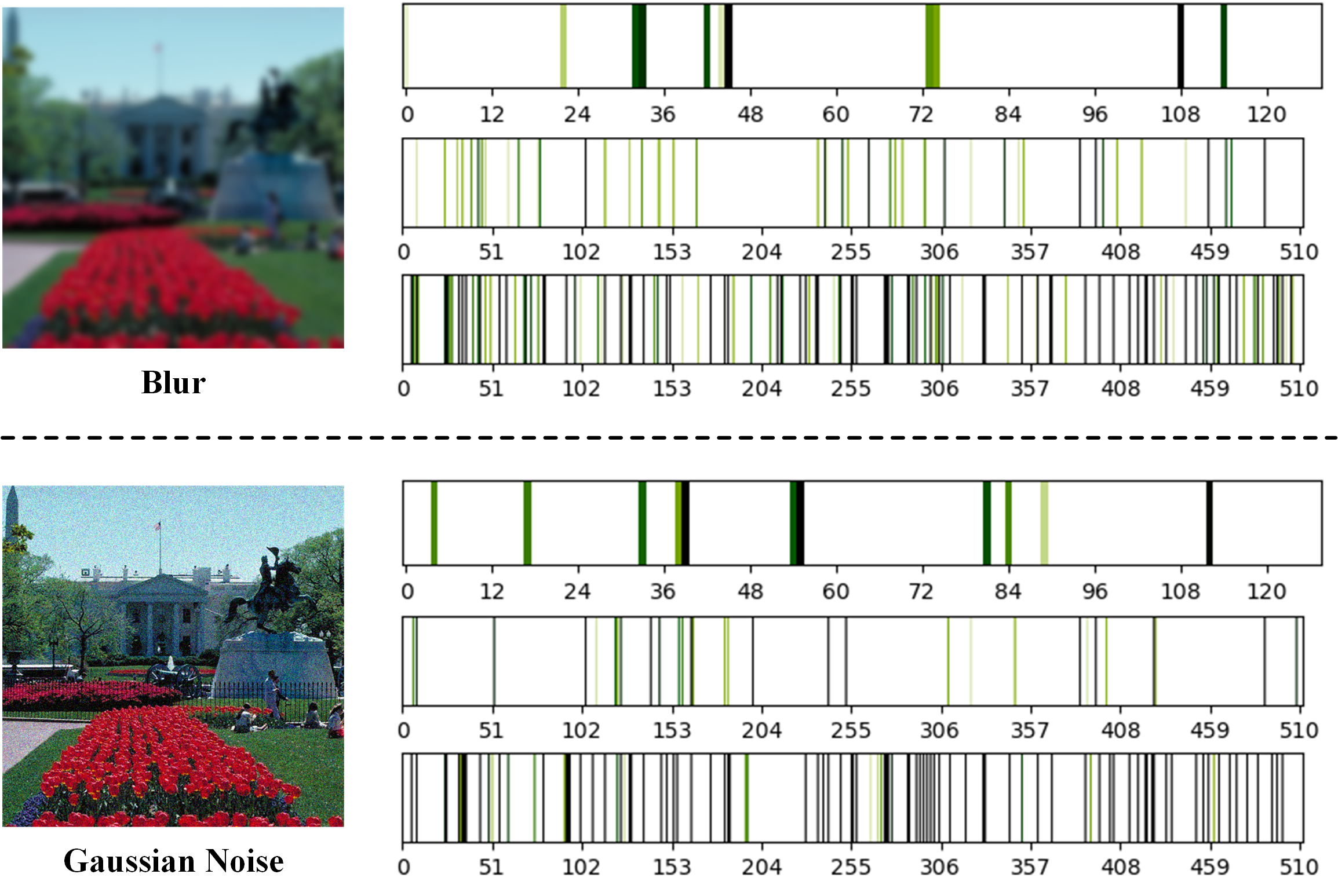}
	\caption{
		Effects of causal interventions on the perceptual representations of images under various distortions.
		The displayed channels, from top to bottom, correspond to the second, fourth, and fifth stages of the VGG-16 architecture.
	}
	\label{fig:intervention}
\end{figure}

\subsection{Qualitative Evaluation}

In this section, we visually demonstrate the effect of causal intervention on different channels.
As shown in \autoref{fig:intervention}, we demonstrate the visualization effects of causal intervention between the same reference image and different distorted images (blurred and additive Gaussian noise) via the Our-VGG architecture.
These visualizations reveal the varying levels of sensitivity each perceptual channel exhibits to different types of distortions in the pretrained model, thus providing a clear indication of how such distortions alter the perceptual characteristics of the image.
This disparity in the nature of distortion impacts highlights the channel-specific nature of perceptual changes.
By isolating the causal effects, we can identify specific features or regions of the image that are most sensitive to certain types of distortions, thus further elucidating how different distortions affect overall image perception.

\section{Conclusion}
In conclusion, we developed an FR-IQA method grounded in abductive counterfactual inference to analyse the causal relationships between deep network features and perceptual distortions.
We investigated the causal effects of deep features on human perception and incorporated causal reasoning with feature comparisons, constructing a model capable of addressing various complex distortions in different IQA scenarios.
The proposed approach demonstrated strong generalizability, as the analysis of perceptual causal correlations was shown to be independent of the backbone architecture, making it adaptable across diverse deep networks.
Additionally, abductive counterfactual experiments validated the effectiveness of the proposed causal framework, confirming the model’s superior perceptual relevance and interpretability of quality scores.
The experimental results reinforced the robustness and competitiveness of the method, which achieved consistent performance across multiple IQA benchmarks.
For future work, we plan to further explore the causal effects in real-world distortions and investigate the application of the proposed causal model in other vision enhancement tasks.

\newpage
{
	\small
	\bibliographystyle{ieeenat_fullname}
	\bibliography{reference}
}


\end{document}